%% file: Delta_p-Newton_final.tex
\documentclass[journal]{IEEEtran}
\usepackage{graphicx}
\usepackage{xcolor}
\usepackage{amsmath,amsfonts}
\usepackage{amssymb}
\usepackage{amscd}
\usepackage{amsthm}
\usepackage{mathtools}
\usepackage{booktabs}
\usepackage{acronym}
\usepackage{algorithm, algpseudocode}
\usepackage{stfloats}
\usepackage{alltt, epsfig, bbm}
\usepackage{xspace}
\usepackage{hyperref}
\usepackage[caption=false]{subfig}
\usepackage{cite}

\usepackage{dsfont}
\usepackage{amssymb}

\newtheorem{definition}{Definition}

\DeclareMathOperator{\sgn}{sgn}
\newcommand{\X}{\mathbf{X}}
\newcommand{\dual}{\lambda}
\newcommand{\algo}{DpN\xspace}
\newcommand{\shortminus}{\scalebox{0.5}[0.5]{\( - \)}}

\usepackage{makeidx}
\graphicspath{{Figures/}}

\usepackage{xr}
\makeatletter
\newcommand*{\addFileDependency}[1]{
\typeout{(#1)}
\@addtofilelist{#1}
\IfFileExists{#1}{}{\typeout{No file #1.}}
}\makeatother

\newcommand\var[1]{\mathop{\textnormal{\slshape #1}}\nolimits}

\newcommand*{\myexternaldocument}[1]{%
\externaldocument{#1}%
\addFileDependency{#1.tex}%
\addFileDependency{#1.aux}%
}

\myexternaldocument{SupMat_final}

\begin{document}
\title{A Newton Method for Hausdorff Approximations of the Pareto Front 
within Multi-objective Evolutionary Algorithms}
\author{Hao Wang, Angel E. Rodriguez-Fernandez, Lourdes Uribe, Andr\'e Deutz, 
        Oziel Cortés-Piña, Oliver Sch\"utze
\thanks{H. Wang and A. Deutz are with the Leiden Institute of Advanced Computer Science of Leiden University, Leiden, The Netherlands.  \\
        A. Rodriguez-Fernandez and O. Sch\"utze are with the Computer Science Department of
        the CINVESTAV, Mexico City, Mexico. L. Uribe is with the Faculty of Actuarial Sciences of Universidad Anáhuac México. O. Cortés-Piña  are with the IPN, Mexico City, Mexico.
        O. Sch\"utze acknowledges support from CONAHCYT 
project CBF2023-2024-1463. }
        }

\markboth{Journal of \LaTeX\ Class Files,~Vol.~14, No.~8, August~2015}%
{Shell \MakeLowercase{\textit{et al.}}: Bare Demo of IEEEtran.cls for IEEE Journals}

\maketitle              

\begin{abstract}
A common goal in evolutionary multi-objective optimization is to 
find suitable finite-size approximations of the Pareto front of a given 
multi-objective optimization problem. While many multi-objective evolutionary algorithms have proven to be very efficient in 
finding good Pareto front approximations, they may need quite a few 
resources or may even fail to obtain optimal or nearly optimal approximations. 
Hereby, optimality is implicitly defined by the chosen performance 
indicator.\\
In this work, we propose a set-based Newton method for Hausdorff approximations of the Pareto front to be used within multi-objective 
evolutionary algorithms. To this end, 
we first generalize the previously proposed Newton step for the performance indicator to treat constrained problems for general reference sets. To approximate the target Pareto front,
we propose a particular strategy for generating the reference set that utilizes the data gathered by the evolutionary algorithm during its run. Finally, we show the benefit of the Newton method as a post-processing step on several benchmark test functions and different base evolutionary algorithms.  

\end{abstract}

\section{Introduction}

\IEEEPARstart{M}{ulti}-objective optimization refers to the concurrent optimization 
of several objectives (\hspace{1sp}\cite{deb:01,Coello07}). One important characteristic of such continuous multi-objective optimization problems (MOPs) is that one can expect 
the solution set -- the Pareto set, respectively, its image, the Pareto front -- to form 
at least locally an object of dimension $k-1$, where $k$ is the number of objectives 
involved in the problem. 
For the numerical treatment of such problems, specialized evolutionary strategies called multi-objective evolutionary algorithms (MOEAs) have caught the interest of 
researchers and practitioners during the last three decades (\hspace{1sp}\cite{sinha:22,interview_deb:23}). 
Algorithms of this type do not depend on the initial conditions, are very robust, and only require a few assumptions on the model. Another advantage is that the population-based
approach allows, in principle, the finite-size representation of the Pareto 
set/front to be obtained in one algorithm run. It is widely accepted and shown in
numerous empirical studies that MOEAs are capable of computing reasonable Pareto 
front approximations in rather short times. However, it is also known that 
they require quite a few resources to obtain very good approximations and that
they may never reach optimal ones (\hspace{1sp}\cite{bringmann:14,schuetze_hernandez:21}). 
Given a multi-objective performance indicator, a Pareto front approximation is 
optimal (w.r.t.~this indicator)
if it yields the best indicator value among all other possible approximations of 
the same magnitude. \\
In this work, we propose a set-based Newton method to be used within MOEAs that aims for Hausdorff approximations of the Pareto front. The method uses the averaged Hausdorff distance $\Delta_p$~\cite{schuetze_deltap:12} as a performance indicator. Since the method requires a target set (ideally the Pareto front), we propose in this 
study a new way to represent the best regions found by the MOEA by utilizing the 
data gathered by this algorithm during the run of the algorithm. The cost of the Newton method depends, among others, on the size of this target set $T$ and the 
relation between $T$ and the Newton iterate. In particular, the Newton step
gets highly simplified if these two sets are ``matched'' (i.e., if there exists a 
one-to-one relationship for the Newton step between the two sets).
To realize this, we will apply a particular subset selection on $T$
and propose a particular matching strategy for the reduced target set and the 
Newton iterate. We will demonstrate the strength of the novel Newton method as a post-processing tool to refine results obtained by the base MOEA. The main 
contributions can be summarized as follows: 

\begin{itemize}
\item  Derivation of the set based Newton steps for  the Generational Distance ($GD$), the Inverted Generational Distance ($IGD$), and $\Delta_p$ for 
  constrained MOPs and for general target sets. 

\item Proposal of a new reference set generator aiming for Pareto front approximations
  by utilizing the data gathered by the MOEA during the algorithm run. 

\item Proposal of a particular Newton method for ``matched sets'', including 
  a particular matching strategy between the Newton iterate and the (reduced)
  Pareto front approximation.  
\end{itemize}

A set-based Newton method for the indicator $\Delta_p$ has been proposed in 
\cite{uribe:20}. This method, however, is restricted to unconstrained problems and
uses target sets that have to be given a priori. The method can be
applied to aspiration set problems as addressed in \cite{rudolph_aspiration:14}. However, it is
unclear how to obtain Pareto front approximations. Further, it is used as a standalone method and may get stuck in local minima. The present study aims to fill all of these gaps. 

The remainder of this work is organized as follows: in Sec. II, we shortly 
recall the required background and discuss the related work. In Sec. III, we derive the Newton steps for constrained MOPs and general target sets. 
The Newton method for matched sets is proposed in Sec. IV. In Sec. V, we 
show its strength as a post-processing tool, and finally, we draw our conclusions
and give possible paths of future work in Sec. VI.

\section{Background and Related Work}

In this work, we consider continuous constrained multi-objective optimization
problems that can be expressed as 

\begin{equation}
\label{eq:MOP-general}
\begin{array}{cl}
\min\limits_{x} & \; F(x),\\
\mbox{s.t.}   & \; h_i(x) = 0,\quad i=1,\ldots, p, \\
              & \; g_i(x) \leq 0,\quad i=1,\ldots, m,
\end{array}
\end{equation}

where $Q\subset \mathbb{R}^n$ is the domain that is defined by the constraint functions, and $F:Q\subset\mathbb{R}^n\to\mathbb{R}^k$ is the map defined by the individual objectives 
$f_i:Q\subset\mathbb{R}^n\to\mathbb{R}$, $i=1,\ldots,k$, 
\begin{equation}      
F:Q\to\mathbb{R}^k,\qquad F(x) = (f_1(x),\ldots,f_k(x))^T. 
\end{equation} 
Optimality is defined by the concept of dominance (e.g., \cite{deb_ehrgott:23}).
The vector 
$v\in\mathbb{R}^k$ is said to be {\em less than} $w\in\mathbb{R}^k$ ($v<_pw$), if $v_i< w_i$ for all 
$i\in\{1,\ldots,k\}$ (analog for $\leq_p$). $y\in Q$ is said to be  {\em dominated by} $x\in Q$ ($x\prec y$) 
w.r.t. problem (\ref{eq:MOP-general}) if $F(x)\leq_p F(y)$ and  $F(x)\neq F(y)$.
$x\in Q$ is called a  {\em Pareto point} 
 or 
  {\em optimal}   if there exists no $y\in Q$ that dominates $x$.
The set $P_Q$ of all Pareto optimal points is termed the {\em Pareto set} and its image $F(P_Q)$  
the {\em Pareto front}.\\
We are in our work interested in the averaged Hausdorff distance $\Delta_p$ (\hspace{1sp}\cite{schuetze_deltap:12,bogoya:20}). 
This distance is a combination of (slight variations of) the Generational Distance (GD, \cite{Veldhuizen99}) and the Inverted Generational Distance (IGD, \cite{Coello05b}). 

\begin{definition}
\label{def:Deltap}
Let $A,B\subset\mathbb{R}^m$ be finite sets. The value
\begin{equation}
\Delta_p(A, B)=\max(GD_p(A,B), IGD_p(A,B)),
\end{equation}
where $p \in \mathbb{N}$ and 
\begin{equation}
 \begin{split}
  GD_p(A,B) &=\left(\frac{1}{|A|} \sum_{a\in A} dist(a,B)^p\right)^{1/p}\\[2mm]
 IGD_p(A,B) &= \left(\frac{1}{|B|} \sum_{b\in B} dist(b,A)^p\right)^{1/p}\\[2mm]
 dist(a,B) &= \underset{b \in B} {\min} \|a - b\|_2,
\end{split}
\end{equation}
is called the {\em averaged Hausdorff distance} between $A$ and $B$.
\end{definition}

For the 
treatment of continuous sets, we refer to~\cite{bogoya:20}. In our study, we are 
particularly interested in measuring the distances between the images of the candidate 
solutions and the Pareto front of a given MOP. \\
Evolutionary multi-objective optimization (EMO) has been a very active
research field for three decades \cite{interview_deb:23}. 
Up to date, many different and powerful multi-objective evolutionary algorithms 
(MOEAs) have been proposed. These algorithms can roughly be divided
into three main classes: (i) algorithms that are based on the dominance
relation to generating pressure toward the Pareto front (e.g., NSGA-II \cite{deb:nsga2} 
and SPEA2 \cite{zlt2002a}), (ii) algorithms that use decompositions of the given 
objective map (e.g., MOEA/D \cite{ZL07} and NSGA-III \cite{nsga3_1}), and (iii) 
algorithms that utilize performance indicators to perform the selection. MOEAs of 
the last class include SMS-EMOA \cite{beume:07}, HypE \cite{bader:11}, and LIBEA \cite{zapotecas:19} for the 
Hypervolume indicator \cite{zitzler2003performance},
R2-EMOA  \cite{trautmann:13} and MOMBI  \cite{raquel:13} for R2 \cite{hansen:98}, and DDE \cite{cynthia:12} for $\Delta_p$. 
An alternative is to use specialized archivers that can be used as external
archives to, in principle, any existing MOEA to obtain certain approximations
w.r.t. a~given indicator~\cite{knowles_lebesque:03,lopez:11, rudolph:16, schuetze_hernandez:21,hernandez_bounded:22}.\\
One way to enhance the overall performance is to hybridize evolutionary algorithms with specialized local search
techniques. This includes the combination with classical mathematical programming techniques such as SQP \cite{gao2008multi}, 
interior point methods \cite{datta2017radial}, direct search methods \cite{zapotecas_cec:11,zapotecas:12}, stochastic local search \cite{shukla2007gradient,biswas:23} and multi-objective continuation methods \cite{schuetze_engopt:08,martin:18,cuate:20}. 
Further, including descent directions is quite prominent among such hybrid MOEAs. The first 
researchers have probably
been Brown and Smith \cite{brownsmith:05} who adopted the steepest descent direction from Fliege and Svaiter \cite{fliege:00}. 
Bosman \cite{bosman2011gradients} provided an analytical description of all descent directions at a given candidate solutions
and combined the resulting descent line search with some MOEAs. Lara et al. proposed a gradient-based hybrid capable of performing movements toward and along the Pareto front \cite{lara_hcs}. Later, specialized local
search operators have been proposed that can be applied with or without gradient information (the latter
via using existing neighborhood information) and that can cope with constraints \cite{schutze2016directed,schutze2017gradient,URIBE2021100938}. 
Finally, set-based approaches exist similar to the one proposed in this study, which considers the 
performance indicator as a particular scalar optimization problem defined in higher dimensional
search space. Emmerich et al. proposed a Hypervolume gradient-based search in \cite{EmmerichD12}. In subsequent studies,
researchers investigated the set-based Newton method for the Hypervolume indicator \cite{sosa:20,wang:23} and for  $\Delta_p$ \cite{uribe:20}.

The subset selection problem consists of selecting a representative subset out of a set 
according to a particular objective, and is an active research topic both in the machine 
learning and EMO field. In particular, in the latter, 
subset selection is, e.g., used (i) to maintain the population size or (ii) to
compare results coming from an unbounded archiver to the final population of a 
MOEA.  According to \cite{Shang_SubSelSummary}, there exist four categories of subset selection: exact algorithms (e.g., \cite{26_GUERREIRO}), greedy algorithms (\hspace{1sp}\cite{SubSel_IB_10},\cite{SubSel_DB_12},\cite{22_Chen}), iterative algorithms such as distance-based (\hspace{1sp}\cite{SubSel_DB_13}) and clustering-based methods (\hspace{1sp}\cite{Ishibuchi_IGD_clustering}), and evolutionary algorithms (\hspace{1sp}\cite{32_Ishibuchi}). 
In particular, for clustering-based subset selection methods, it was shown that the objective function of $k$-means and IGD coincide (\hspace{1sp}\cite{uribe:20,Ishibuchi_IGD_clustering}), making it the natural subset selection choice for this work since the main contribution for the $\Delta_p$ indicator is $IGD$ for the case when solutions are close to the Pareto front.

\section{The $\Delta_p$-Newton Step for Constrained MOPs}
Here, we develop the population-based $\Delta_p$-Newton step for constrained MOPs.  To this end, we will first 
separately discuss the respective steps for $GD$ and $IGD$ using
general reference sets $Z$. Next, we will discuss the $\Delta_p$-Newton
step for the special case of {\em matched sets} leading to significant
simplifications. 
In the following discussions, we shall always take $p=2$ in the $\Delta_p$ indicator. 
Throughout this section, we assume $Z$ will be given. We will discuss
the complete Newton method in the next section, including the
generation of suitable (matched) reference sets. 

\subsection{The $GD$-Newton Step}

\paragraph{Handling equalities}
We consider in the following the problem
\begin{equation}
\label{eq:MOP}
\begin{array}{cl}
\min\limits_{x} & \; F(x),\\
\mbox{s.t.}   & \; h(x) = 0, 
\end{array}
\end{equation}

\noindent where $h(x) = (h_1(x),\ldots, h_p(x))^\top$ and  $h_i:\mathbb{R}^n \to \mathbb{R}$, $i=1,\ldots, p$. That is,
$Q$ is defined by $p$ equality constraints. \\
Given a magnitude $\mu\in\mathbb{N}$, we are considering the following set based optimization problem

\begin{equation}
\label{eq:GDmin}
 \max_{X\subset Q\atop |X| = \mu} GD_2^2(X),
\end{equation}

\noindent where $GD_2^2(X)$ denotes $GD_2(X,Z)^2$ for 
a given set  $X = \{x^{(1)},\ldots, x^{(\mu)}\}$ of candidate solutions 
$x^{(i)}\in\mathbb{R}^n$, $i=1,\ldots, \mu$, that measures the distance between $F(X)$ and a given approximation
$Z = \{z_1,\ldots,z_M\}\subset\mathbb{R}^k$ of the Pareto front of the considered MOP (the latter is assumed to be given
here, we address the computation of $Z$ in the next subsection).  Note that we have, for our purpose, changed the input of 
GD compared to Definition \ref{def:Deltap} since $X$ is the only variable in our setting.\\
We stress that considering a set $X\subset \mathbb{R}^n$ of magnitude $\mu$
is essentially the same as considering a point $\X$ in $\mathbb{R}^{\mu n}$ (to see this, consider e.g. \\
$\X = (x^{(1)}_1,\ldots, x^{(1)}_n, x^{(2)}_1,\ldots, x^{(2)}_n, \ldots, x^{(\mu)}_1,\ldots, x^{(\mu)}_n) 
 \in \mathbb{R}^{\mu n}$). Consequently,
problem (\ref{eq:GDmin}) can be viewed as a ``classical'' single-objective optimization problem (SOP) of dimension $\mu n$
of the decision variable space.
We will, in the following, consider (\ref{eq:GDmin}) as such an SOP to derive its KKT equations, which are 
needed to define the Newton step. \\
Feasibility of all elements $x^{(j)}$ of the set $\X$ means
\begin{equation}
 h_i(x^{(j)}) = 0,\qquad  i = 1,\ldots, p, \; j = 1,\ldots, \mu. 
\end{equation}

For the related SOP defined on $\mathbf{X}$ we hence establish 

\begin{equation}
h_{i,j}:\mathbb{R}^{\mu n} \to \mathbb{R},\qquad h_{i,j}(\X) = h_i(x^{(j)}),
\end{equation}

\noindent for  $i\in\{1,\ldots,p\}$ and $j\in\{1,\ldots, \mu\}$, as well as the constraint 
function $\bar{h}:\mathbb{R}^{\mu n}\to\mathbb{R}^{p \mu}$ via 

\begin{equation}
\bar{h}(\X) = 
   \begin{pmatrix} h_{1,1}(\X)\\h_{2,1}(\X)\\\vdots\\h_{p,1}(\X)\\h_{1,2}(\X)\\h_{2,2}(\X)\\\vdots\\h_{p,2}(\X)\\\vdots\\h_{1,\mu}(\X)\\\vdots \\  h_{p,\mu}(\X)
   \end{pmatrix}
   \coloneqq
   \begin{pmatrix} \bar{h}_1(\X)\\ \bar{h}_2(\X)\\ \vdots\\ \bar{h}_p(\X)\\ \bar{h}_{p+1}(\X)\\ \bar{h}_{p+2}(\X)\\\vdots\\\bar{h}_{2p}(\X)\\\vdots\\\bar{h}_{(\mu-1)p+1}(\X)\\\vdots\\\bar{h}_{\mu p }(\X) 
   \end{pmatrix}.
\end{equation}

The derivative of $\bar{h}$ at $\X$ is given by 

\begin{equation}
 \bar{H} \coloneqq \nabla \bar{h}(\X) = \operatorname{diag}\left( H(x^{(1)}), \ldots, H(x^{(\mu)})\right) \in \mathbb{R}^{\mu p \times \mu n},
\end{equation}

where $H(x^{(i)})$ denotes the derivative of $h$ at $x^{(i)}$, 

\begin{equation}
 H(x^{(i)}) = \begin{pmatrix} \nabla h_1(x^{(i)})^\top\\ \vdots \\ \nabla h_p (x^{(i)})^\top\end{pmatrix}
           \in \mathbb{R}^{p\times n}. 
\end{equation}

We are now in the position to write down the Karush-Kuhn-Tucker (KKT) equations of the SOP related to (\ref{eq:GDmin}):

\begin{equation}
\label{eq:KKT}
\begin{split}
\nabla GD_2^2(\X) + \bar{H}^\top\dual &= 0\\
\bar{h}(\X)                            &= 0,
\end{split}
\end{equation}

\noindent where $\lambda \in\mathbb{R}^{\mu p}$ is a Lagrange multiplier (see supplementary material for the definition of the derivatives of $GD$). Finding KKT points of 
the SOP related to (\ref{eq:GDmin})
is hence equivalent to the root-finding problem `$R_{GD}(\X,\lambda) = 0$', where 

\begin{equation}
\label{eq:R_GD=0}
\begin{split}
R_{GD}\colon& \mathbb{R}^{\mu (n + p)} \to \mathbb{R}^{\mu (n + p)}\\
R_{GD}(\X,\lambda) &= \begin{pmatrix} \nabla GD_2^2(\X) + \bar{H}^\top\lambda \\\bar{h}(\X) \end{pmatrix},
\end{split}
\end{equation}
and where $\X \in\mathbb{R}^{\mu n}$ and $\lambda \in\mathbb{R}^{\mu p}$. The derivative of $R_{GD}$ at $(\X,\lambda)^T$ 
is given by 

\begin{equation}
 DR_{GD}(\X,\dual) = \begin{pmatrix} \nabla^2 GD_2^2(\X) + S & \bar{H}^\top\\ \bar{H} & 0 \end{pmatrix}
 \in \mathbb{R}^{\mu(n+p)\times \mu(n+p)},
\end{equation}
where
\begin{equation}
 S = \sum_{j=1}^{\mu p} \lambda_j \nabla^2 \bar{h}_j(\X)\in\mathbb{R}^{\mu n \times \mu n}. 
\end{equation}

We are now in the position to state the set-based Newton step for $GD_2^2$: 
given $\X_k\in\mathbb{R}^{\mu n}$ and $\lambda_k \in\mathbb{R}^{\mu p}$, 
the new iterate $(\X_{k+1},\lambda_{k+1})$ solves the system 

\begin{equation}
\label{eq:Newton_step_S}
 DR_{GD}(\X_k,\dual_k)
\begin{pmatrix} \X_{k+1} - \X_{k}\\ \dual_{k+1} - \dual_{k}\end{pmatrix} 
  = - R_{GD}(\X_k,\dual_k). 
\end{equation}

The Newton step is well defined if $DR_{GD}(\X_k,\dual_k)$
is regular. This is, e.g., the case if $\bar{H}$ has full
row rank and if $\nabla^2 GD_2^2(\X) + S$ is positive definite
on the tangent space of the constraints (\hspace{1sp}\cite{nocedal:06}).

For problems with linear constraints, the 
computational cost reduces significantly: 
since $S$ in $DR_{GD}$ vanishes we obtain the new iterate via solving 
\color{black}
\begin{equation}
\begin{split}
\label{eq:newton-step_full}
  \begin{pmatrix} \nabla^2 GD_2^2(\X_k) & \bar{H}^\top \\ \bar{H} & 0
  \end{pmatrix}
 \begin{pmatrix} \X_{k+1} - \X_k \\ \dual_{k+1} - \dual_k
  \end{pmatrix}  = \\[2mm]
  - 
 \begin{pmatrix} \nabla GD_2^2(\X_k) + \bar{H}^\top \dual_k \\ \bar{h}(\X_k)
  \end{pmatrix}. 
  \end{split}
\end{equation}

In this case, the block structure 
of the  matrix in (\ref{eq:newton-step_full})
-- induced by the block structures of $\nabla^2 GD_2^2(\X_k)$ and $\bar{H}$ -- can be used to reduce the computational
complexity. If we denote $$\Delta\X_k \coloneqq \X_{k+1} - \X_k =: (\Delta x_k^{(1)},\ldots, \Delta x_k^{(\mu)})^\top,$$ 
$$\Delta \lambda_k \coloneqq \dual_{k+1} - \dual_k =: (\Delta \lambda_k^{(1)},\ldots, \Delta \lambda_k^{(\mu)})^\top,$$ and 
$\lambda^{(i)} \coloneqq (\lambda_{(i-1)p+1},\ldots, \lambda_{ip})\in\mathbb{R}^p$, $i=1,\ldots, \mu$, then each
tuple $(\Delta x_k^{(i)},\Delta \lambda_k^{(i)})$, $i=1,\ldots,\mu$, can be computed via solving the system

\begin{equation}
\begin{split}
\label{eq:newton-step_indiv}
\begin{pmatrix} \frac{2}{\mu} Dg_{GD}(x_k^{(i)}) & H(x_k^{(i)})^\top \\ H(x_k^{(i)}) & 0
  \end{pmatrix} 
\begin{pmatrix} \Delta x_k^{(i)} \\ \Delta \lambda_k^{(i)}
  \end{pmatrix}
 = \\[2mm]
 \begin{pmatrix} -\frac{2}{\mu}J(x_k^{(i)})^\top (F(x_k^{(i)})-z_{j_i}) - H(x_k^{(i)})^\top \lambda_k^{(i)}\\ 
  -h(x_k^{(i)})
  \end{pmatrix},
  \end{split}
\end{equation}

where 

\begin{equation}
 Dg_{GD}(x_k^{(i)}) = J(x_k^{(i)})^\top J(x_k^{(i)}) + \sum_{l=1}^k \alpha_l \nabla^2 f_l(x_k^{(i)})\in\mathbb{R}^{n\times n},
\end{equation}
and where $\alpha\in\mathbb{R}^k$ is as defined in Eq. (9) (suppl. mat.). The 
computation of $\X_{k+1}$ and $\lambda_{k+1}$ hence reduces from $\mathcal{O}(\mu^3 (n+p)^3)$ -- if computed via 
(\ref{eq:newton-step_full}) using classical Gaussian Elimination -- down to $\mathcal{O}(\mu (n+p)^3)$ when using
(\ref{eq:newton-step_indiv}) for $i=1,\ldots, \mu$. This is the case since (\ref{eq:newton-step_full}) is
a system of linear equations of size $\mu(n+p)\times \mu(n+p)$, and (\ref{eq:newton-step_indiv}) represents $\mu$ systems
of size $(n+p)\times (n+p)$. 
Hence, via using the block structure, the computation cost then is, in particular, linear 
in size $\mu$ of $\X$ instead of cubic. \\
The use of the block structure further enables individual step sizes for each $(\Delta x_k^{(i)},\Delta \lambda_k^{(i)})$
which we have done for the computations presented in this work. More precisely, we have used the initial
step size $t_0 = 1$ together with backtracking and Armijo's condition (e.g., \cite{nocedal:06}). 

For the results presented below, we have used
the Newton step (Eq.~\eqref{eq:newton-step_full}), 
i.e., we have omitted $S$ for MOPs with nonlinear constraints. While omitting $S$ leads to relatively slow (linear) and non-Newton-like convergence, it comes, on the other hand, with the benefits described above (reduction of the computational cost, step size control). 
For problems with highly nonlinear constraints,
however, it may be advisable to use 
(Eq.~\eqref{eq:Newton_step_S}) for the Newton step. 

\paragraph{Handling inequalities}

For the computations presented in the sequel, we have adopted the following simple strategy
(\hspace{1sp}\cite{beltran:20}): all inequalities that are nearly active at $\X$ where a step into the search direction 
would lead to no improvement of these constraints are treated as equalities at $\X$. All
other inequalities are  (locally) disregarded. 
See supplementary material for more details. In doing so, the Newton 
step is computed as described above. This simple strategy turned 
out to be  very effective, in particular  
since we are hybridizing the search with evolutionary strategies and 
hence only compute rather short trajectories.  It may be interesting
-- in particular concerning a convergence analysis of the 
method -- to consider more sophisticated constraint handling
techniques (as e.g. discussed in \cite{nocedal:06}), which we
leave, however, for future work.

\subsection{The $IGD$-Newton Step}
The $IGD$-Newton step is derived analogously to the $GD$ step, see supplementary
material for details.

\subsection{The $\Delta_p$-Newton Step for Matched Sets}
\begin{figure}
    \centering
    \includegraphics[width=0.75\linewidth]{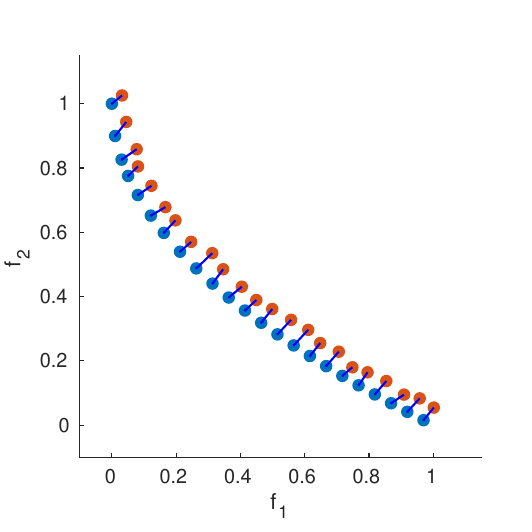}
    \caption{Hypothetical example where the sets $Z$ (blue dots) and 
      $\X$ (the image $F(\X)$ in red dots) are matched. The matching
      is indicated by the lines.}
    \label{fig:matching}
\end{figure}

The $\Delta_p$-Newton step for general reference sets $Z$ is now clear 
by the definition of $\Delta_p$ and the above discussion (see supplementary 
material for details). In the following, we discuss the Newton step for the special case 
where the two sets $\X$ and $Z$ are ``matched"  (we will discuss a particular 
matching strategy in the next section). The use of matched sets significantly simplifies 
the Newton method  -- $GD$ and $IGD$ steps coincide in this case -- and we have experienced an 
improvement in the numerical performance of the method. \\
We say that the population $\X=\{x^{(1)},\ldots,x^{(\mu)}\}$ and the reference set $Z$ are matched if $|Z| = \mu$ and if the set of closest elements $z_{j_i}$
is equal to $Z$ (i.e., $\{z_{j_1},\ldots,z_{j_\mu}\} = Z$). 
In other words, there exists a one-to-one relationship for the
Newton step between the two sets (see also Figure \ref{fig:matching}). For this case, 
the individual steps of $GD$ (see Eq. (\ref{eq:newton-step_indiv}))
and $IGD$ (see Eq. (\ref{eq:IGD_newton-step_indiv}) of the suppl. mat.) get simplified: for $i\in \{1,\ldots,
\mu\}$ we get $m_i = 1$ and $I_i = \{i\}$. 
Using $z_{j_i} = z_i$ (after possible reordering of $Z$) and
$\mu=M=1$ (due to the match), we obtain 

\begin{equation}
\begin{split}
\label{eq:Delta_p_step_matching}
\begin{pmatrix} 2 Dg(x_k^{(i)}) & H(x_k^{(i)})^\top \\ H(x_k^{(i)}) & 0
  \end{pmatrix} 
\begin{pmatrix} \Delta x_k^{(i)} \\ \Delta \lambda_k^{(i)}
  \end{pmatrix}
 = &\\[2mm]
 \begin{pmatrix} -2J(x_k^{(i)})^\top (F(x_k^{(i)})-z_i) - H(x_k^{(i)})^\top \lambda_k^{(i)}\\ 
  -h(x_k^{(i)})
  \end{pmatrix},
  \end{split}
\end{equation}

where 

\begin{equation}
 Dg(x_k^{(i)}) = J(x_k^{(i)})^\top J(x_k^{(i)}) + \sum_{l=1}^k (f_l(x^{(i)})-z_{i,l})\nabla^2 f_l(x_k^{(i)}),
\end{equation}

for both the $GD$ and the $IGD$ step. 

\section{The $\Delta_p$-Newton Method within MOEAs}

Here, we propose a particular realization of the $\Delta_p$-Newton method, which is facilitated 
by the use of the data coming from the run of an evolutionary algorithm applied to the given 
MOP.\\
Of foremost importance is certainly the proper choice of the reference set $Z$ since
the Pareto front of the given problem is not known a priori. Further, we will
need an initial iterate $\X_0$ for the Newton method and a suitable direction $\eta$ to further
shift elements of $Z$ into infeasible regions if necessary. Algorithm \ref{alg:Deltap-NewtonMethod:summary}
shows the four main steps of our proposed method: 
\begin{enumerate}
    \item[A.] Merge and clean existing populations
    \item[B.] Compute the first iterate $(X_0,\lambda_0)$ 
    \item[C.] Generate the reference set $Z$ 
    \item[D.] Apply the $\Delta_p$-Newton method 
\end{enumerate}

We will describe each of the four steps in more detail in the following subsections and will finally present the proposed Newton method in step D (Algorithm \ref{alg:Newton}).

\begin{algorithm}[!t]
\caption{$\Delta_p$-Newton Method}
\begin{algorithmic}[1]
\renewcommand{\algorithmicrequire}{\textbf{Input:}}
\renewcommand{\algorithmicensure}{\textbf{Output:}}
\Require MOEA populations $P_f,\dots, P_{f\shortminus(\kappa\shortminus 1)s}$, number of iterations $N_i$.
\Ensure final iterate $(\X_{N_i},\lambda_{N_i})$
\State $P \leftarrow$ merge and clean $P_f, \dots, P_{f\shortminus(\kappa\shortminus 1)s}$
\State $(\X_0,\lambda_0) \gets$ compute the first iterate from $P$
\State $Z \gets$ generate the reference set from $P$ 
\State $(X_{N_{i}},\lambda_{N_{i}}) \gets$ apply $\Delta_p$-Newton Method using $(\X_0,\lambda_0)$
\State \Return $\{(\X_{N_i},\lambda_{N_i})\}$
\end{algorithmic}
\label{alg:Deltap-NewtonMethod:summary}
\end{algorithm}

\subsection{Merge and Clean}

Denote by $P_i$ the population of the $i$-th generation computed by the MOEA, and let $P_f$ be the final 
population ($f$ being the number of the last generation). As the basis for our considerations, we use
\begin{equation}
\label{eq:P-prime}
 P' = P_f\cup P_{f\shortminus s}\cup \ldots \cup P_{f\shortminus(\kappa\shortminus 1)s}, 
\end{equation}
i.e., the union of $\kappa$ of the last populations, using a gap of $s$ generations. The latter is done to gather additional information on the Pareto front rather than just using $P_f$. First, we 
remove unpromising candidates from $P'$. These are, in particular, (i) points that are not nearly feasible 
and (ii) potential outliers, i.e., points that are non-dominated within $P'$, but those images may 
be far away from the Pareto front (such points may be computed by the MOEA in the presence of
weakly optimal solutions that are not globally optimal, \cite{ikeda:01}). Elements $p\in P'$ as described in (ii) can be eliminated using the auxiliary objectives proposed in 
\cite{ishibuchi:20}:
\begin{equation}
    \label{eq:bar_f}
     \bar{f}_i(x) = (1-\omega)f_i(x) + \frac{\omega}{k}\sum_{i=1}^k f_i(x),\quad i=1,\ldots, k, 
\end{equation}
where $\omega>0$ is ``small'' (we have used $\omega = 0.02$). 
This can be realized using a non-dominance test on $P'$ with the auxiliary objectives. 
Alternatively (and recommended), one can use these objectives directly in the run of the MOEA.
The removal of these unpromising solutions leads to the subset $P$. 
If $|P|\leq 0.1 \mu$, we do not apply the Newton method since, in this case, applying the costly local search does not seem promising. Applying the Newton method, we always
use $\mu$ elements. See supplementary material for the case that $|P|<\mu$.\\

\subsection{Compute the first iterate $(\X_0,\lambda_0)$}

Next, we obtain the set $\X_0 = \{x_0^{(1)},\ldots, x_0^{(\mu)}\}$ from $P$ via $k$-medoids. 
This is done due to the relation of $k$-means and the $IGD$-indicator reported in  (\hspace{1sp}\cite{uribe:20,Ishibuchi_IGD_clustering}). However, since we require the decision variables that $k$-means does not provide, we use $k$-medoids instead. Finally, we set all the
elements of $\lambda_0$ to $0$, mainly because these are the values one obtains if the optimal solution 
$\X^*$ is located in the interior of the domain $Q$. A summary of this process can be seen in Algorithm \ref{alg:ComputingFirstIterate} together with the merge and clean step (in lines 1 and 2).

\begin{algorithm}[!t]
\caption{Computing $(\X_0,\lambda_0)$}
\begin{algorithmic}[1]
\renewcommand{\algorithmicrequire}{\textbf{Input:}}
\renewcommand{\algorithmicensure}{\textbf{Output:}}
\Require set of solutions $P$, population size $\mu$.
\Ensure initial set $(\X_0,\lambda_0)$
\State $P' \leftarrow P_f\cup \ldots \cup P_{f\shortminus(\kappa\shortminus 1)s}$
\State $P \gets $ remove unpromising elements from $P'$
\State $\X_0 \coloneqq \{x_0^{(1)},\ldots, x_0^{(\mu)}\} \gets k$-medoids($P,\mu$)
\State $\lambda_0 \leftarrow (0,\ldots,0)$
\State \Return $(\X_0,\lambda_0)$
\end{algorithmic}
\label{alg:ComputingFirstIterate}
\end{algorithm}

\subsection{Generate $Z$}

The process of obtaining the reference set $Z$ from the MOEA populations involves six intermediate steps:
\begin{enumerate}
    \item[1.] Component detection
    \item[2.] Filling
    \item[3.] Generation of the unshifted reference set
    \item[4.] Computation of the shifting direction
    \item[5.] Shifting
    \item[6.] Matching
\end{enumerate}
The overall aim is to find a reference set of size $\mu$ that is evenly spread along the PF (or at least along
the parts of the front that have been detected so far). This set is then shifted downwards to allow the Newton 
method to detect better solutions. To this end, we first need to detect the different connected 
components (Step 1), which is done using DBSCAN \cite{DBSCAN}. Next, in Step 2, each of the components is filled to obtain a gap-free approximation of each component. The effect and importance of this step are shown in the 
supplementary material. After that (Step 3), we reduce the size of the filled set down to size $\mu$ using $k$-means 
(due to its relation to $IGD$), leading to $T$. This set is shifted toward the utopian region. For this, a 
shifting direction $\eta$ for each component is computed that is ideally orthogonal to this set (Step 4). Then, in 
Step 5, $T$ is shifted in this direction, leading to $Z$. 
Finally, a matching between $Z$ and the first iterate $\X_0$ is done. Each of these steps is described in detail 
in the supplementary material. 

\subsection{The $\Delta_p$-Newton Method}

We have now collected all the elements for the population-based $\Delta_p$-Newton method 
we propose in the following.\\ 
Algorithm \ref{alg:Newton} shows the pseudo-code of the Newton method. First, $P'$, 
$P$, $Z_0 \coloneqq Z$, and $\eta$ are computed as described above.  The method terminates if the magnitude of $P$  does not exceed 10 percent of the magnitude of $P'$. This is due to the 
fact that too few promising solutions may not be enough to build up a suitable reference set. 
Otherwise, the Newton method is performed. Each iteration consists of two steps: the Newton step 
and the update of the reference set. If one element $z_i$ is detected to 
be reachable, i.e., if a candidate solution $x_{k+1}^{(i)}$ is computed, those image are 
approximately given by $z_k^{(i)}$, then this target is further shifted by $t\eta$ so that 
the subsequent Newton iterates aim for better solutions. 

\begin{algorithm}[!t]
\caption{$\Delta_p$-Newton Method}
\begin{algorithmic}[1]
\renewcommand{\algorithmicrequire}{\textbf{Input:}}
\renewcommand{\algorithmicensure}{\textbf{Output:}}
\Require initial iterate $(\X_0,\lambda_0)$, step $t>0$,
tolerance $tol_y>0$, merged 
sets $P, P'$, number of iterations $N_i$, reference set $Z$,
shifting direction $\eta$
\Ensure {final iterate $(\X_{N_i},\lambda_{N_i})$}
\State $Z_0 \leftarrow Z$
\If {$|P| \leq 0.1 |P'|$}
\State \textbf{terminate} \Comment{Do not apply Newton method}
\EndIf
\For {$l=0,1,\ldots,N_i-1$}
  \State compute $(\X_{l+1},\lambda_{l+1})$ via (\ref{eq:Delta_p_step_matching})
\For {$i=1,\ldots,\mu$} \Comment{Update $Z$}
\If {$\|F(x_{k+1}^{(i)})-z_k^{(i)}\|_2 < \var{tol}_y$}
  \State $z_{k+1}^{(i)} \leftarrow z_k^{(i)} + t\eta$
\Else
  \State $z_{k+1}^{(i)} \leftarrow z_k^{(i)}$
\EndIf
\EndFor
\EndFor
\\ \Return{$(\X_{N_i},\lambda_{N_i})$}
\end{algorithmic}
\label{alg:Newton}
\end{algorithm}

\begin{figure} 
    \centering\captionsetup{width=.45\linewidth}
  \subfloat[$F(P_1)$]{%
       \includegraphics[width=0.5\linewidth]{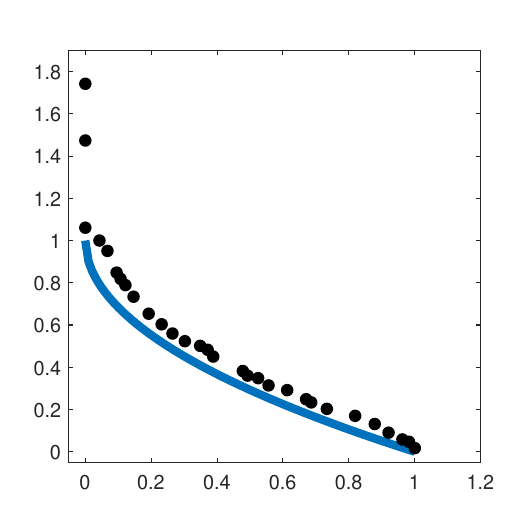}
       \label{subfig:zdt1Pop1}
       }
    \subfloat[$F(P_2)$]{%
       \includegraphics[width=0.5\linewidth]{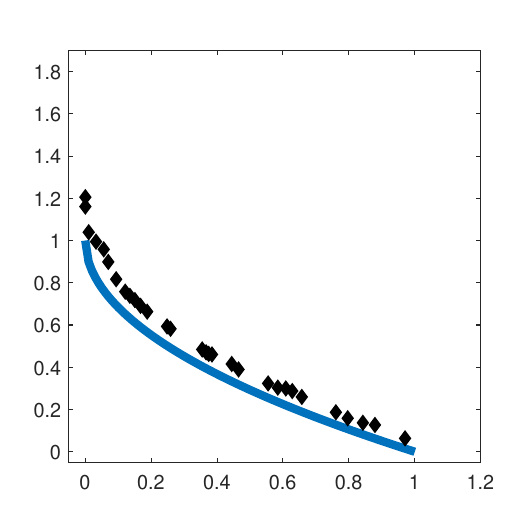}
       \label{subfig:zdt1Pop2}
       }
  \\
  \subfloat[$Y$ (in red), T (magenta diamonds), Removed Outliers (black crosses)]{%
        \includegraphics[width=0.5\linewidth]{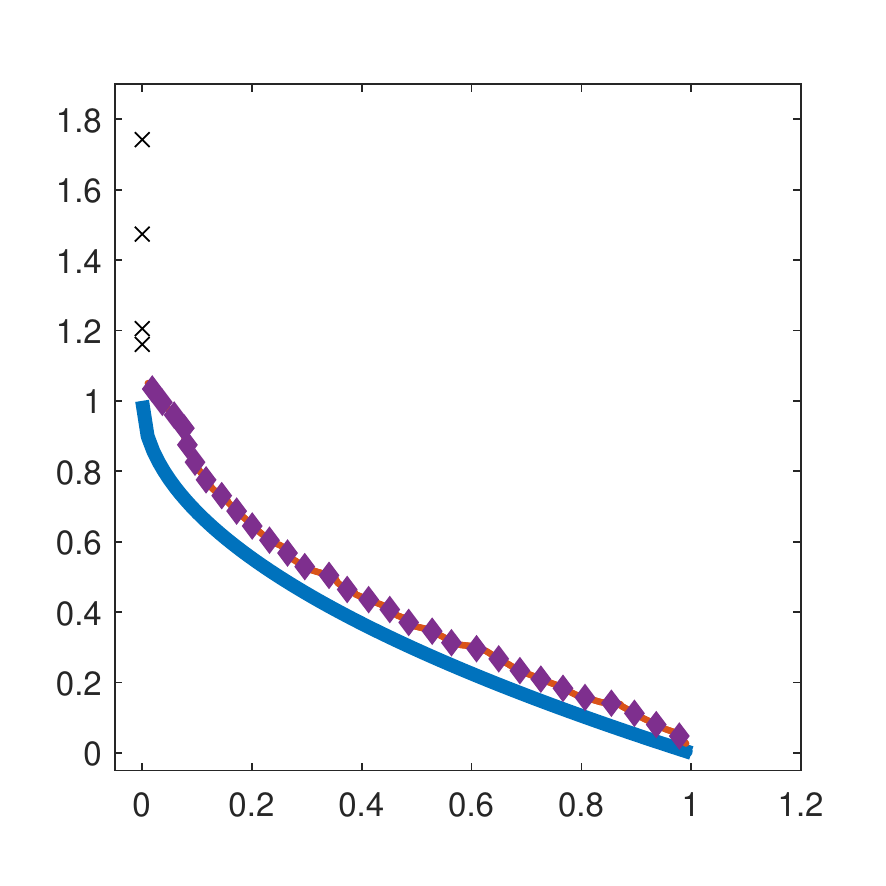}
        \label{subfig:zdt1filling}
        }
  \subfloat[Matching of $\X_0$ (black dots) with $Z$ (magenta diamonds)]{%
        \includegraphics[width=0.5\linewidth]{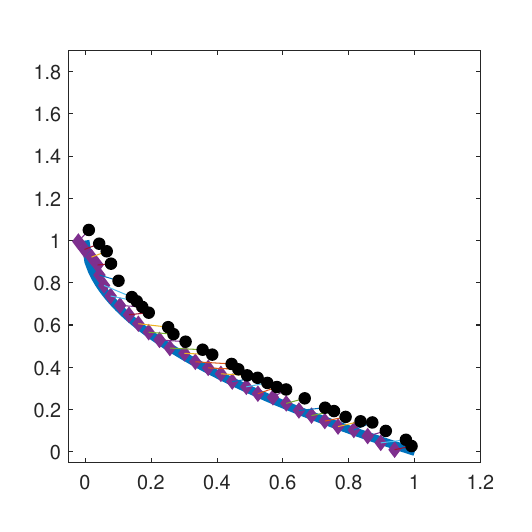}
        \label{subfig:zdt1shifting}
        }
  \\
  \subfloat[Newton]{%
        \includegraphics[width=0.5\linewidth]{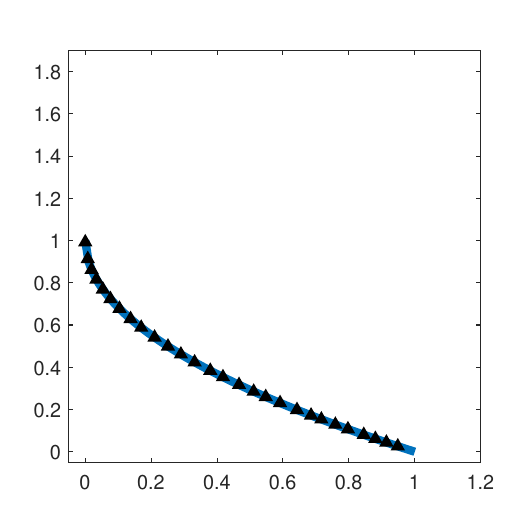}
        \label{subfig:zdt1newtonresult}
        }
  \subfloat[Left axis: IGD, right: $||R(X)||$]{%
        \includegraphics[width=0.535\linewidth, trim=1mm 0mm 9mm 12mm, clip]{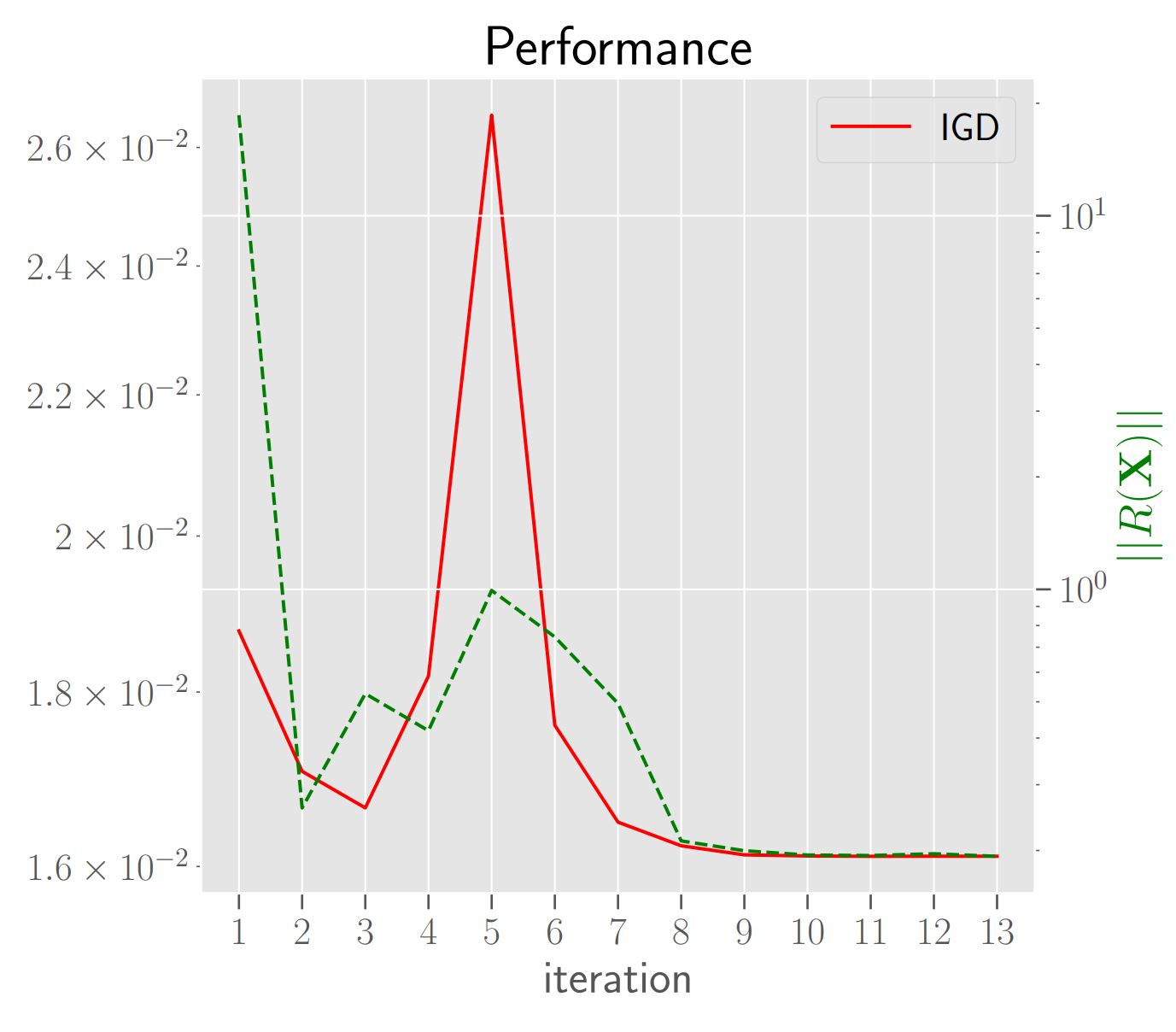}
        \label{subfig:zdt1performance}
        }
  \caption{Application of the $\Delta_p$-Newton method on ZDT1.  The starting populations are shown in (a) and (b), the filled set ($Y$) and the unshifted reference set ($T$) are shown in (c), the matching between the initial iterate $\X_0$ and the shifted reference set $Z$ can be seen in (d) and finally in (e) the result of the Newton method can be seen together with its performance in (f).}
  \label{fig:zdt1} 
\end{figure}

Figure \ref{fig:zdt1} shows an application of the Newton method on ZDT1 using
$n=3$ decision variables and $k=2$ objectives. Figures \ref{fig:zdt1} (a)  and (b) 
show the Pareto front together with two hypothetical populations $P_f$ and $P_{f-s}$, 
respectively, using population size $\mu=30$. Both populations contain outliers that 
are not contained in $P'$. Figure \ref{fig:zdt1} (c) shows the resulting set $Y$ (red dots) 
and the initial targets $T$ (magenta diamonds). The black crosses are the detected
outliers. Figure \ref{fig:zdt1} (d)
shows the image of $\X_0$ together with the matching with the shifted set $Z$ for the step
size $t=0.05$. Figure \ref{fig:zdt1} (e) shows the final result of the Newton method using 
$N_i = 6$.  Apparently, the Newton method could significantly improve the overall approximation 
quality, for instance, by reducing gaps in the representation. This observation is supported by the $\Delta_2$ 
values of the two approximations: it is $\Delta_2(F(P_f),F(P_Q)) = 1.0170$ and 
$\Delta_2(F(\X_6),F(P_Q)) = 0.1701$. 

\begin{figure} 
    \centering\captionsetup{width=.45\linewidth}
  \subfloat[$F(P_1)$]{%
       \includegraphics[width=0.5\linewidth]{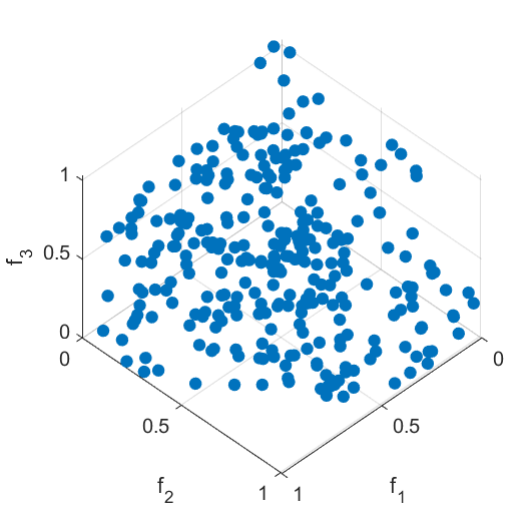}}
    \subfloat[$F(P_2)$]{%
       \includegraphics[width=0.5\linewidth]{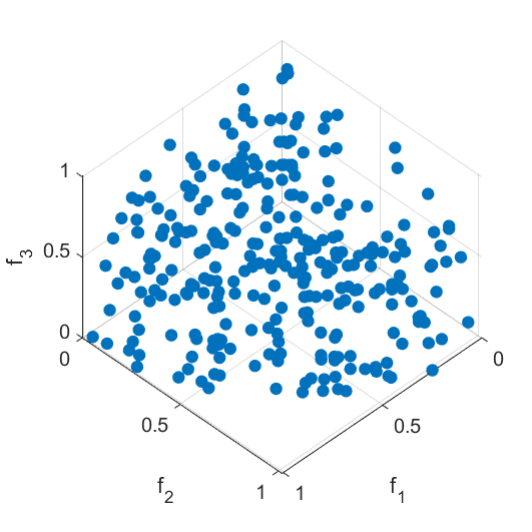}}
  \\
  \subfloat[$Y$ (in blue), T (magenta diamonds).]{%
        \includegraphics[width=0.5\linewidth]{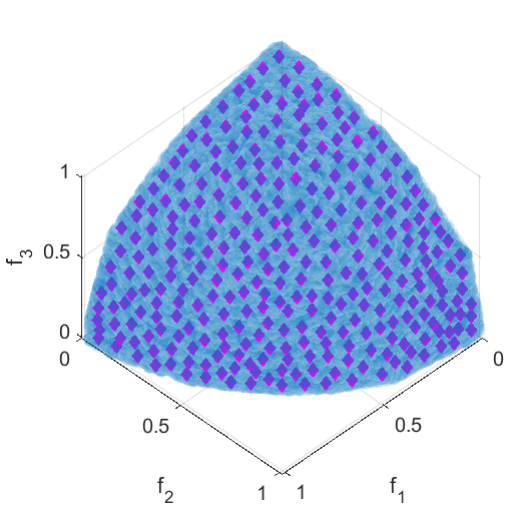}}
  \subfloat[Matching of $\X_0$ (black dots) with $Z$ (magenta diamonds)]{%
        \includegraphics[width=0.5\linewidth]{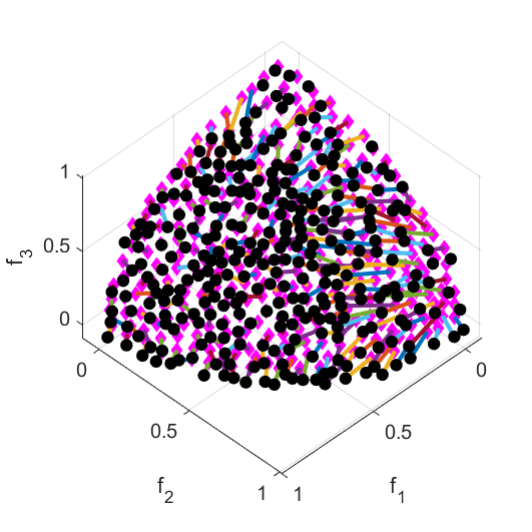}}
  \\
  \subfloat[Newton]{%
        \includegraphics[width=0.5\linewidth]{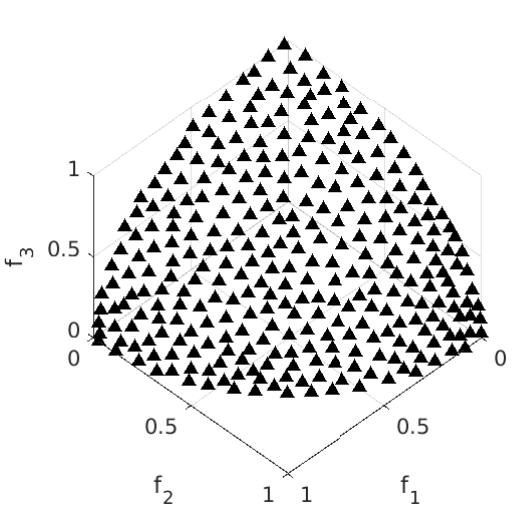}}
  \subfloat[Left axis: IGD, right: $||R(X)||$]{%
        \includegraphics[width=0.56\linewidth, trim=13mm 3mm 15.5mm 18mm, clip]{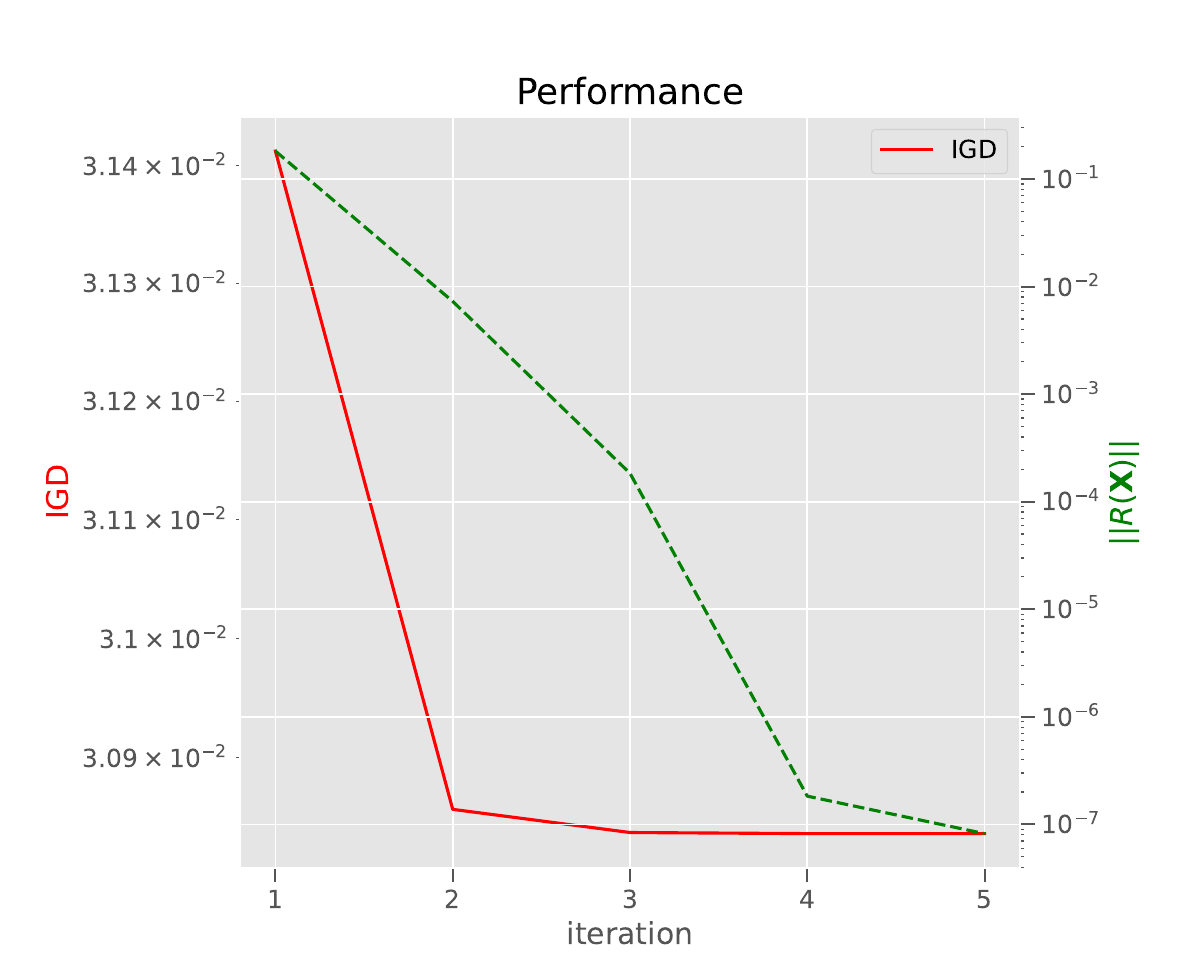}}
  \caption{Application of the $\Delta_p$-Newton method on DTLZ2.  Two starting populations are shown in (a) and (b), the filled set ($Y$) and the unshifted reference set ($T$) are shown in (c), and the matching between the initial iterate $\X_0$ and the shifted reference set $Z$ can be seen in (d) and finally in (e) the result of the Newton method can be seen together with its performance in (f).}
  \label{fig:DTLZ2_EA} 
\end{figure}

Figure \ref{fig:DTLZ2_EA} shows the application of the Newton method on DTLZ2 ($n=12$, $k=3$)  using populations obtained with NSGA-II. Similarly, Figure \ref{fig:DTLZ2_EA} (a) shows the last generation ($P_{300}$) of NSGA-II and Figure \ref{fig:DTLZ2_EA} (b) shows $P_{295}$. We use $\kappa = 4$ generations for problems with $k\geq 3$, but we only show those two. Figures \ref{fig:DTLZ2_EA} (c), (d), (e), and (f) show the same sets as in Figure \ref{fig:zdt1} but without the PF.

\subsection{Complexity Analysis} 

We now present the complexity of the whole process (Algorithm \ref{alg:Deltap-NewtonMethod:summary}) for connected Pareto fronts. If the PF is disconnected, the complexity is the same as described below, but for each component, the total size is replaced by each component size accordingly.

The complexity of the entire method is  governed by generating the unshifted reference set,  the matching, and the $\Delta_p$-Newton method,  and is 
\begin{equation}
    \mathcal{O}(\tau_2 k\mu N_f + \mu^2\log\mu + \mu (n+p)^3),
\end{equation}
 where $\tau_2$ is the number of iterations of $k$-means, $k$ is the number of objectives, $\mu$ is the population size, $N_f$ is the size of the filling, $n$ is the number of decision variables, $p$ is the number of equality constraints and we assume that the size of the merged and cleaned populations ($\ell$)  is proportional to  $\mu$ and $k\ll\mu$. 
 The complexity analysis of each individual step is discussed in the supplementary material. 

\section{Numerical results}\label{sec:results}
In this section, we empirically show the strength of the novel approach. Due to the relatively high cost of the Newton method, it does not
seem advisable to use it as an operator within a MOEA that is applied with a certain
probability in each step. Instead, it seems to be more reasonable to use it as a
post-processing step after the run of the chosen base MOEA. \\
We first discuss the effect of the Newton method similar to Figure \ref{fig:zdt1} (see
all figures and further discussion in the supplementary material). To this end,
we have chosen to use next to ZDT1 ($k=2$ objectives, connected and convex Pareto
front) also ZDT3 ($k=2$, disconnected convex-concave PF), DTLZ1 ($k=3$, connected
linear PF), DTLZ2 ($k=3$, connected concave PF), DTLZ7 ($k=3$, disconnected convex-concave
PF), and CONV4-2F ($k=4$, disconnected convex PF). These functions have been selected
to show the method's capability to handle MOPs with different PF shapes. Figures 
\ref{fig:ZDT1_EA} to \ref{fig:CONV4-2F_EA_123} of the supplementary material show ``before-after''  approximations together with some intermediate steps
of the reference set generation on selected (median) results. Table \ref{tab:numerical-results} shows the
$\Delta_p$ values of the obtained approximations for these problems averaged over
30 independent runs. As can be seen by the indicator values and visually, the Newton method 
significantly increases the approximation qualities for all chosen test problems. 
Since, again, the Newton method comes with a certain cost, the natural question that
arises is what the base MOEA would have achieved using these additional
resources. To address this issue, we have chosen 
to use the frequently used and highly proven MOEAs  (1) NSGA-II~\cite{deb:nsga2}; 
(2) NSGA-III~\cite{nsga3_1,JainD14}; (3) MOEA/D~\cite{Rubio-LargoZV15} and
(4) SMS-EMOA~\cite{beume:07} as base algorithms. For each MOEA, we set the population size to $\mu = 100$ for bi-objective problems and $\mu=300$ 
for three-objective ones.  For the implementation of MOEAs, we use the \texttt{pymoo} library (version 0.6.1.1).  We take the following well-applied test problems: 
\begin{itemize}
    \item ZDT1, ZDT2, ZDT3, ZDT4, and ZDT6~\cite{DebS06a}: $k=2$, decision 
    space $[0,1]^{30}$ for ZDT1-3 and $[0,1]^{10}$ for ZDT4 and 6. 
    \item DTLZ1-7~\cite{DebTLZ05}: $k=3$, decision space $[0,1]^{10}$ 
    for DTLZ2-7 and $[0,1]^7$ for DTLZ1.
    \item IDTLZ1-4~\cite{JainD14}: $k=3$, decision space $[0,1]^{11}$.
    \item CF1-7 ($k=2$) and CF8-9 ($k=3$)~\cite{zhang2008multiobjective}:
    all functions are, in addition to the box constraints, restricted by
    a further nonlinear inequality constraint. The dimension of the decision space is $n=10$.
    \item CONV4-2F (defined in the supplementary material): $k=4$, decision space $[-3,3]^4$. %
\end{itemize}
\begin{figure*}[!t]
\centering
\includegraphics[width=0.95\linewidth, trim=2mm 7mm 0mm 0mm, clip]{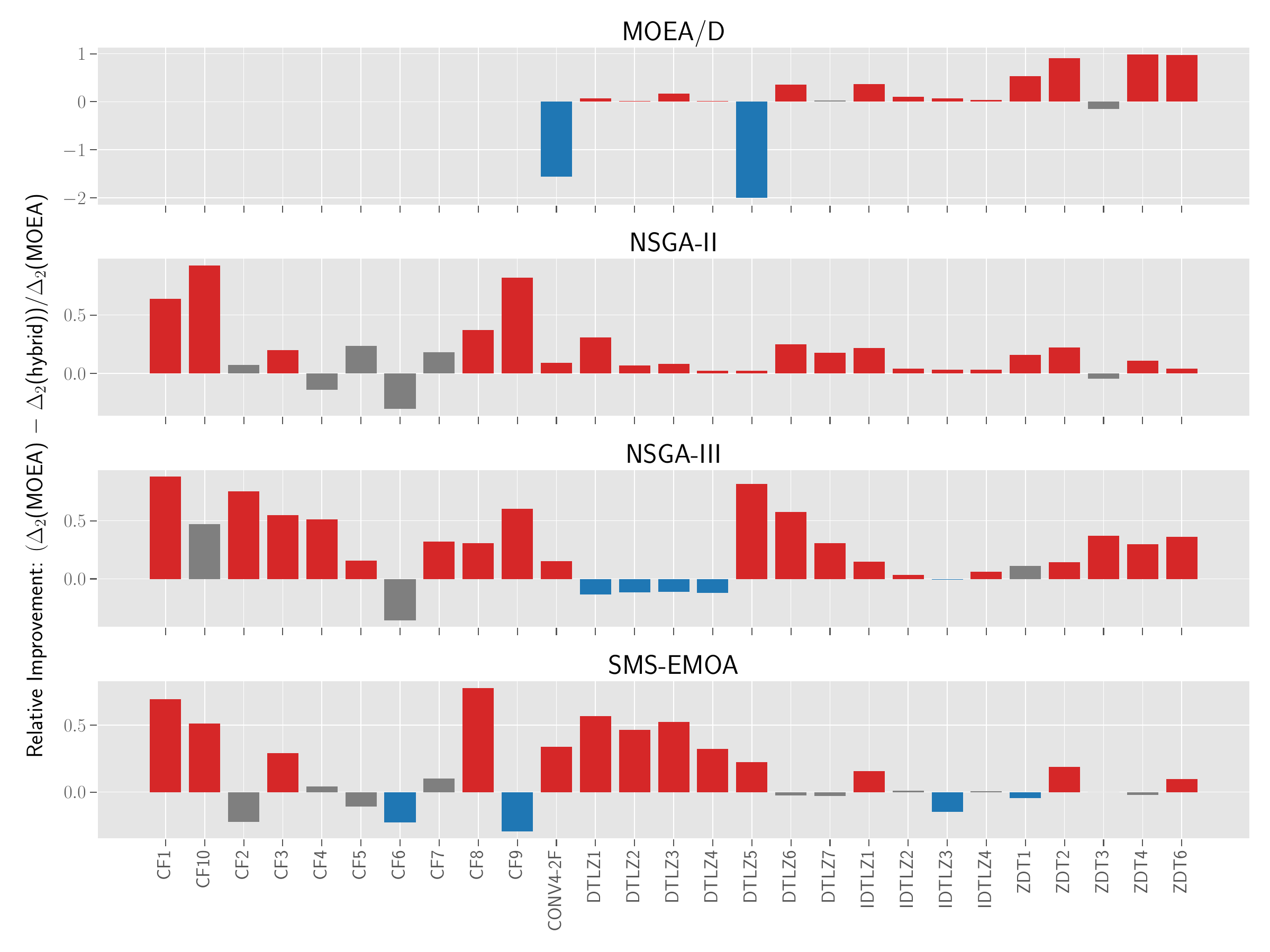}
  \caption{Results on performance comparison of the hybrid method (MOEA + \algo) and MOEA. We show the relative improvement of the hybrid method over the MOEAs in terms of the median $\Delta_2$ values on each problem, i.e., $(\Delta_2$(MOEA) $-$ $\Delta_2$(hybrid))$/\Delta_2$(MOEA), where red bars indicate statistically significant improvement according to Mann-Whitney U test; blue ones indicate significant worsening; gray ones show no statistical difference. 30 independent runs are conducted for each method on each problem to obtain the result.}
  \label{fig:relative-improvement} 
\end{figure*}

For all problems, at least a part of the Pareto set is located at the boundary of $Q$. Thus, the constraint-handling techniques described above are required for all problems.  For all base MOEAs except MODA/D\footnote{Since the constraint-handling method is not implemented in the \texttt{pymoo} library, we did not test the MOEA/D algorithm on all CF problems.}, we handle the constraints 
using the adaptive $\varepsilon$-constraint method~\cite{TakahamaS06}. This method considers a solution 
feasible subject to a small threshold $\varepsilon$, which decreases linearly to zero. The initial value of $\varepsilon$ is set to the average constraint value of the initial population. In our experiment, we diminish $\varepsilon$ to zero after 50\% of the iterations of the MOEA. 

In our experiments, we execute the MOEA for 300 iterations and then run 
 the Newton method (in short: \algo)  for 6 iterations. We used 
automatic differentiation (AD)~\cite{0002126} techniques to compute the Jacobian and Hessian of the objective 
functions for \algo. Theoretically, the AD's time complexity to compute the Jacobian is always upper-bounded by $4\operatorname{OPS}(f)$ in the reverse mode~\cite{Margossian19}, where $\operatorname{OPS}(f)$ denotes the total number of additions and multiplications required in one function evaluation. Note that this upper bound holds for arbitrary real-analytic functions. To make the comparison more realistic, we empirically measured the prefactor in this upper bound over all problem tested, yielding an upper bound of $1.836\operatorname{OPS}(f)$. Also, we perform the same estimation for the time complexity bound of the Hessian, which is $3\operatorname{OPS}(f)$. As it is the empirical worst-case scenario among all tested problems, we think it is reasonable and does not bring artificial benefits to the Newton method on individual functions. During the 6 iterations of the \algo, we record the actual number of AD calls (which varies from one run to another due to the backtracking line search) to compute the Jacobian and Hessian for each run on each problem. Taking on the upper bound of AD's time complexity, we calculate the number of the FEs equivalent to the AD's computation in 6 iterations of \algo. Finally, we run the MOEA again with a budget of 300 generations + additional FEs equivalent to the AD calls to make the comparison fair.\footnote{The source code of the Newton method and data can be accessed at \url{https://github.com/wangronin/HypervolumeDerivatives}.}

\begin{figure}
\centering
\includegraphics[width=0.95\linewidth, trim=2mm 7mm 0mm 0mm, clip]{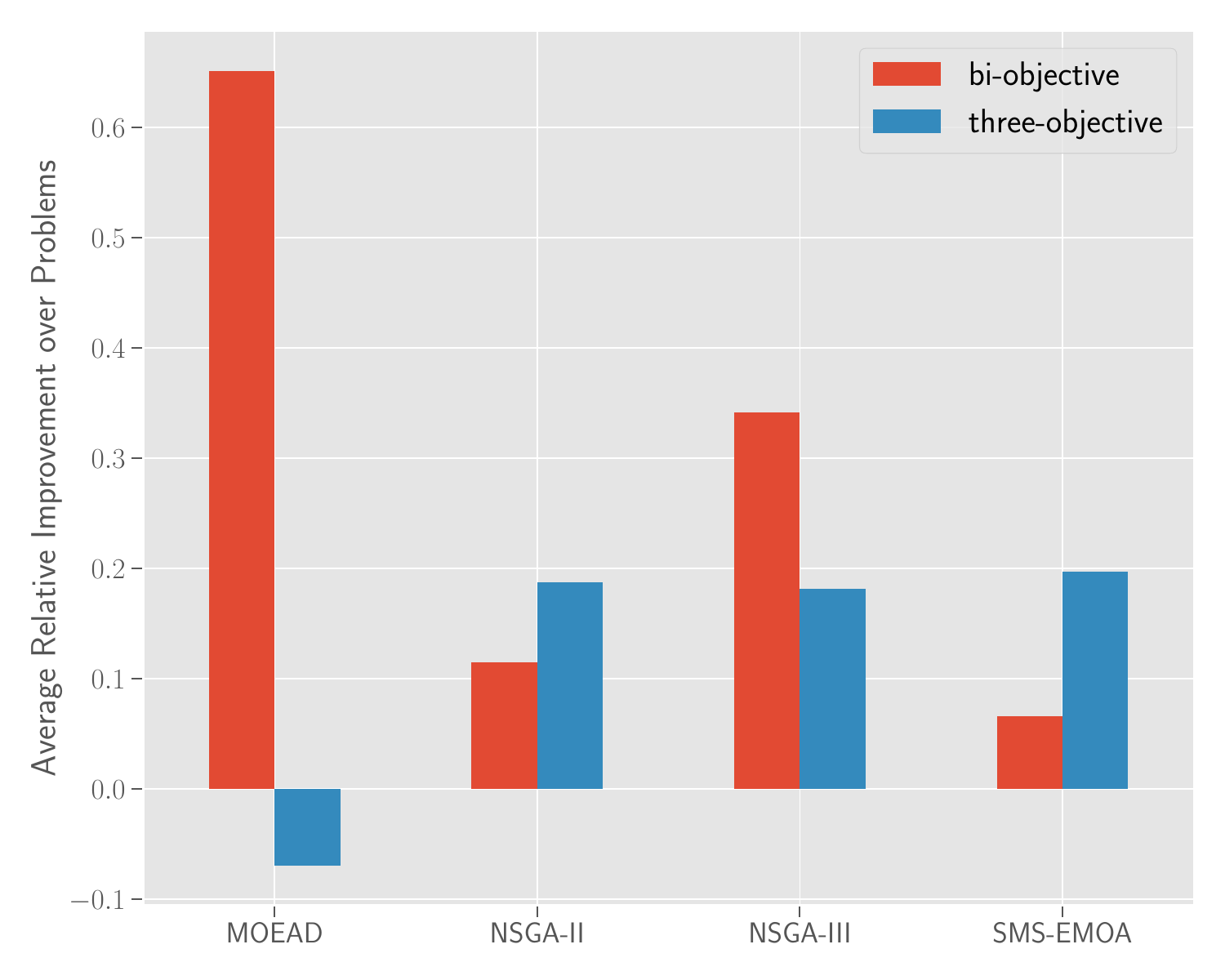}
  \caption{We show the impact of the number of objectives by aggregating the relative improvement for bi-objective and tri-objective problems, respectively. Since we only consider one four-objective problem (CONV4-2F) in this study, this problem is not shown in this chart. }
  \label{fig:improvement-aggregated} 
\end{figure}

Figure~\ref{fig:relative-improvement} and  Table~\ref{tab:numerical-results} contain the main results: we compute  the medians of $\Delta_p$ ($p=2$) 
between the final approximation set of each algorithm and the Pareto front, estimated from 30 independent runs 
for each algorithm-problem pair. The Mann-Whitney U test (with a 5\% significance level) is used to check 
statistical significance.  Figure~\ref{fig:relative-improvement} shows the relative improvement of the hybrid method (MOEA + $\Delta_p$ Newton) over the MOEAs in terms of the median $\Delta_2$ values on each problem, i.e., $(\Delta_2$(MOEA) $-$ $\Delta_2$(hybrid))$/\Delta_2$(MOEA). Table~\ref{tab:numerical-results} shows the detailed $\Delta_2$ values.  Among 98 algorithm-problem pairs, the hybridization of MOEA and \algo 
outperforms the standalone MOEA on 66 cases, ties on 21 cases, and loses on only 11 cases (66/21/11). 
Considering the base MOEAs separately, we obtain 21/6/0 for NSGA-II, 19/3/5 for NSGA-III, 13/10/4 for
SMS-EMOA and 13/2/2 for MOEA/D. As mentioned before, we did not test MOEA/D on CF problems since in \texttt{pymoo}, the constraint-handling method is not implemented for MOEA/D.  Furthermore, in Figure~\ref{fig:improvement-aggregated}, we depict the relative improvement of the hybrid method over the bi-objective and tri-objective problems to investigate the impact of the dimension of objective space on the performance: (1) when using MOEA/D as the initial algorithm in the hybrid, the performance degenerates quickly as the dimension increases; (2) the performance of the hybrid method improves quite a bit in higher dimensions for SMS-EMOA; (3) for NSGA-II, the performance does not differ much across dimensions; (4) for NSGA-III, the performance decays a bit on tri-objective problems. \\
We further examine the cases where \algo loses: it is mainly attributed to either a nearly perfect reference set or a poor one. For the former, we see on DTLZ1-4, the approximation sets of NSGA-III are nearly perfect; hence, \algo can not improve much upon it. We observe the same behavior on ZDT1 with SMS-EMOA. The latter results from either bad convergence to or coverage of the Pareto front by the MOEAs. For IDTLZ3, neither NSGA-III nor SMS-EMOA manages to get sufficiently close to the Pareto front, and the reference set interpolation does not help much since \algo can easily get stuck in local efficient sets. For CF6 and CF9, SMS-EMOA only covers a small fraction of the Pareto front upon termination. Hence, the reference set we generate also has poor coverage of the Pareto front. For DTLZ5, MOEA/D only covers a fraction of the Pareto front, similar to the previous case. For CONV4-2F, we find out that the shifting direction $\eta$ computed from  MOEA/D's final population is not pointing towards the Pareto front, causing the \algo to move further away.

Based on this result, we conclude that overall \algo accelerates the empirical convergence compared to the genetic operators in MOEA when the approximation set is already decently close to the Pareto set. 

\input{results-gen=300}

\section{Conclusions and Future Work}
In this paper, we have proposed a new set-based Newton method aiming for Hausdorff approximations
of the Pareto front of a given multi-objective optimization problem (MOP). The method was designed
to be used within multi-objective evolutionary algorithms (MOEAs). The basis for this
method has been the $\Delta_p$-Newton method proposed in \cite{uribe:20}, for which we have 
applied three crucial modifications: (i) the Newton steps for the $GD$, the $IGD$, and the $\Delta_p$
indicator have been derived to treat constrained MOPs and general reference
sets. (ii) To aim for Pareto front approximations, we have proposed a particular reference
set generator, for which the data gathered by the used base MOEA is utilized. (iii) To 
significantly reduce the cost of the method, we have considered the Newton method for ``matched sets'' and have proposed a particular matching strategy between the (reduced) reference set and the 
Newton iterate. We have finally shown the strength of the novel method as a post-processing tool
for which we have used four benchmark suites (ZDT, DTLZ, IDTLZ, and CF) and four MOEAs as base
algorithms (NSGA-II, NSGA-III, MOEA/D, and SMS-EMOA).\\
Though the results shown in this work are very promising, some issues may be worth
investigating to further improve the Newton method. One issue is certainly the cost of the method
which results from the use of derivative information. It is conceivable that the use of quasi-Newton
elements or the direct use of neighborhood information, as done in \cite{schutze2016directed,martin:18,URIBE2021100938}, can reduce the cost
of the local search. Another issue is that the outcome of the Newton method highly depends on the 
quality of the reference set. It is worth investigating if the reference set generator can
be improved, e.g., via an update during the Newton iteration or mechanisms that detect (local) degenerations of the solution set. 

\bibliographystyle{plain}
\bibliography{references.bib}

\appendix

\section{Derivatives of $GD$ and $IGD$}
We consider the following continuous multi-objective optimization problem~(MOP)
\begin{eqnarray} \label{eq:mop_suppl}
        \min_{x \in Q} F(x).
\end{eqnarray}

Let $X = \{x^{(1)},\ldots,x^{(\mu)}\} \subset\mathbb{R}^n$ be a candidate set for Eq.~(\ref{eq:mop_suppl})
 and  $Z = \{z_1,\ldots,z_M\}\subset\mathbb{R}^k$ be a given reference set. 
The indicator $GD_p$ measures the averaged distance of the image of $X$ and $Z$: 

\begin{equation}\label{eq:gdp}
GD_{p}(X):= \left( \frac{1}{\mu} \sum_{i=1}^{\mu} d(F(x^{(i)}),Z)^{p}\right)^{\frac{1}{p}}.
\end{equation}

Hereby, we have used the notation 
\begin{equation}\label{eq:gdp_d}
d(F(x^{(i)}),Z):= \min_{j=1, \ldots, M} \Vert F(x^{(i)})- z_{j} \Vert, \;\; \mbox{for} \;\; i=1, \ldots, \mu, 
\end{equation}
and assume $Z$ to be fixed for the given problem (hence, it does not appear as an input argument).  

\subsection{Derivatives of $GD_2^2$}

\paragraph{Gradient of $GD_2^2$}
In the following, we have to assume that for every image $F(x^{(i)})$ there exists 
exactly one element in $Z$ that is nearest to this point. That is, 
$\forall \;\; i=1, \ldots,\mu$ there exists an index $j_{i} \in \lbrace 1, \ldots , M\rbrace$ such that:
 
\begin{equation}
\begin{split}
\label{eq:gdp_assump}
d(F(x^{(i)}),Z) = \Vert F(x^{(i)}) -z_{j_{i}} \Vert  &< \Vert F(x^{(i)}) -z_{q} \Vert  \\[2mm]
& \forall q\in \lbrace 1, \ldots ,M \rbrace\setminus \{j_{i}\}. 
\end{split}
\end{equation}

Otherwise, the gradient of $GD_{p}$ is not defined at  $X$. 
If condition (\ref{eq:gdp_assump})
is satisfied, then (\ref{eq:gdp}) can be written as follows:

\begin{equation}\label{eq:gdpt}
GD_{p}(X):= \left( \frac{1}{\mu} \sum_{i=1}^{\mu} \Vert F(x^{(i)}) -z_{j_{i}} \Vert^{p}\right)^{\frac{1}{p}},
\end{equation}

\noindent and for the special case $p=2$, we obtain 

\begin{equation}\label{eq:gdp_2}
GD_{2}^{2}(X):= \frac{1}{\mu} \sum_{i=1}^{\mu} \Vert F(x^{(i)}) -z_{j_{i}} \Vert_{2}^{2} \in \mathbb{R}^{n \cdot \mu}.
\end{equation}

The gradient of $GD_2^2$ at $X$ is hence given by 
 
\begin{equation}\label{eq:ggd2}
\nabla GD_{2}^{2}(A):= \frac{2}{\mu}\left( \begin{array}{c} 
J(x^{(1)})^{T} (F(x^{(1)})-z_{j_{1}}) \\
J(x^{(2)})^{T} (F(x^{(2)})-z_{j_{2}}) \\
\vdots \\
J(x^{(\mu)})^{T} (F(x^{(\mu)})-z_{j_{\mu}}) \end{array} \right) \in \mathbb{R}^{n \cdot \mu},
\end{equation}

\noindent where $J(x^{(i)})$ denotes the Jacobian matrix of $F$ at $x^{(i)}$ for 
$i=1, \ldots, \mu$. We call the vector

\begin{equation}
J(x^{(i)})^{T} (F(x^{(i)})-z_{j_{i}}),\qquad i\in\{1,\ldots, \mu\},
\end{equation}
the $i$-th {\em sub-gradient} of $GD_2^2$ with respect to $x^{(i)} \in X$.\\

\paragraph{Hessian of $GD_2^2$} 
We first define the map $g: \mathbb{R}^{n} \rightarrow \mathbb{R}^{n}$ as 

\begin{equation}\label{eq:gdp_g}
g(x^{(i)}):= \sum_{l=1}^{k} \alpha_{l}^{(i)} \nabla f_{l}(x^{(i)}),
\end{equation}

\noindent where $\alpha_{l}^{(i)}=: f_{l}(x^{(i)}) -(z_{j_{i}})_l.$ To find an expression of the Hessian matrix, we now derive Eq.~(\ref{eq:gdp_g}) as follows:

\begin{equation}\label{eq:gdp_auxh}
\begin{split}
\mathcal{D}g(x^{(i)}) & = \sum_{l=1}^{k} \left( \nabla f_{l}(x^{(i)})\nabla f_{l}(x^{(i)})^{T} + \alpha_{l} \nabla^{2}f_{l}(x^{(i)})\right) \\
& =J(x^{(i)})^{T}J(x^{(i)})+ W_{\alpha}(x^{(i)}) \in \mathbb{R}^{n \times n},
\end{split}
\end{equation}

\noindent where 
\begin{equation}\label{eq:WalphaGD}
W_{\alpha}(x^{(i)}) = \sum_{l=1}^{k} \alpha_{l} \nabla^{2} f_{l}(x^{(i)}).
\end{equation}

\noindent Thus, the Hessian matrix of $GD^{2}_{2}$ is 

\begin{equation}\label{eq:hgd2}
\nabla^{2} GD^{2}_{2}(X) = \frac{2}{\mu} \mbox{diag} \left(\mathcal{D}g(x^{(1)}),\ldots, \mathcal{D}g(x^{(\mu)})\right) \in \mathbb{R}^{n\cdot \mu \times n \cdot \mu},
\end{equation}

\noindent which is a block diagonal matrix.

\subsection{Derivatives of $IGD_2^2$}
The indicator $IGD_p$ computes how far, on average, the discrete reference set $Z$ is 
from $F(X)$, and is defined as 

\begin{equation}\label{eq:igdp}
IGD_{p}(X):= \left( \frac{1}{M} \sum_{i=1}^{M} d(z_{i},F(X))^{p}\right)^{\frac{1}{p}},
\end{equation}

\noindent where $d(z_{i},F(X))$ is given by

\begin{equation}\label{eq:igdp_d} 
d(z_{i}, F(X)):= \min_{j=1, \ldots, \mu} \Vert z_{i}-F(x^{(j)}) \Vert, \;\; \mbox{for} \;\; i=1, \ldots, M.
\end{equation}

\paragraph{Gradient of $IGD_2^2$}
Similar to $GD$, we will also have to assume that 
$\forall \;\; i=1, \ldots,M$ there exists an index $j_{i} \in \lbrace 1, \ldots , \mu\rbrace$ 
such that:
 
\begin{equation}\label{eq:igdp_assump}
\begin{split}
d(z_{i}, F(X)) = \Vert z_{i}-F(x^{(j)}_{i})\Vert  &< \Vert z_{i} - F(x^{(q)})\Vert  
\\[2mm]
\forall q\in \lbrace 1, \ldots ,\mu \rbrace\setminus \{j_{i}\},
\end{split}
\end{equation}

\noindent since otherwise the gradient of $IGD_{p}$ is not defined. Then, using Eqs.~(\ref{eq:igdp_assump}) and (\ref{eq:igdp}) can be written as follows:

\begin{equation}\label{eq:igdpt}
IGD_{p}(X):= \left( \frac{1}{M} \sum_{i=1}^{M} \Vert z_{i} - F(x^{(j)}_{i})\Vert^{p}\right)^{\frac{1}{p}}.
\end{equation}

From now on we will consider $IGD_{2}^{2}$ which is given by 

\begin{equation}\label{eq:igdp_2}
IGD_{2}^{2}(X):= \frac{1}{M} \sum_{i=1}^{M} \Vert z_{i}- F(x^{(j)}_{i}) \Vert_{2}^{2}.
\end{equation}

In order to derive  the gradient of $IGD_{2}^{2}$, let $I_{l}:=\lbrace i: j_{i} = l\rbrace$, $l\in\{1,\ldots,\mu\}$, be the set formed by the indexes $i \in \lbrace 1, 2, \ldots, M \rbrace$ that are related to $j_{i}$. In other words, this set gives us the relation of the elements of $Z$ related to each image $F(x^{(l)})$. Then, the sub-gradient of $IGD_{2}^{2}$ at point $x^{(l)}$ is given by 

\begin{equation}\label{eq:part_der_igd}
\begin{split}
\frac{\partial IGD_{2}^{2}}{\partial x^{(l)}}(X) &= \frac{2}{M} \sum_{i \in I_{l}} J(x^{(l)})^{T} (F(x^{(l)}) -z_{i})\\[2mm] 
 & = \frac{2}{M} J(x^{(l)})^{T}(m_l F(x^{(l)}) - \sum_{i \in I_{l}} z_{i}),
\end{split}
\end{equation}

\noindent where $m_l = \mid I_{l} \mid.$ Finally, the gradient of $IGD_{2}^{2}$ can be expressed as 

\begin{equation}\label{eq:gigd2}
\nabla IGD_{2}^{2}(X):= \left( \begin{array}{c} 
\frac{\partial IGD_{2}^{2}}{\partial x^{(1)}}(X)\\[2mm]
\frac{\partial IGD_{2}^{2}}{\partial x^{(2)}}(X)\\[2mm]
\vdots \\[2mm]
\frac{\partial IGD_{2}^{2}}{\partial x^{(\mu)}}(X) \end{array} \right) \in \mathbb{R}^{n \cdot \mu}.
\end{equation}

\paragraph{Hessian matrix of $IGD_{p}$}

Analog to the derivation of $GD_{p}-$ Hessian, we first define the map 
$g: \mathbb{R}^{n} \rightarrow \mathbb{R}^{n}$ as 

\begin{equation}\label{eq:igdp_g}
g(x^{(l)}):= J(x^{(l)})^{T} (m_lF(x^{(l)}) - \sum_{i \in I_{l}} z_{i}).
\end{equation}

\noindent Now, let $\sum_{i \in I_{l}} z_{i} = y =\left(y_1,\ldots, y_{k}\right)^{T}$. Then 

\begin{equation}\label{eq:igdp_gy}
\begin{split}
g(x^{(l)}) &= J(x^{(l)})^{T} (m_l F(x^{(l)}) - y) \\[2mm]
 & = m_l \sum_{i=1}^{k} f_{i}(x) \nabla f_{i}(x) - \sum_{i=1}^{k} y_{i}\nabla f_{i}(x).
\end{split}
\end{equation}

\noindent Then, we derive Eq.~(\ref{eq:igdp_gy}) as follows:

\begin{eqnarray}\label{eq:igdp_auxh}
\mathcal{D}g(x^{(l)}) &=&  m_l \sum_{i=1}^{k} f_{i}(x^{(l)}) \nabla^{2}f_{i}(x^{(l)}) + m_l J(x^{(l)})^{T}J(x^{(l)})\\
&& -\sum_{i=1}^{k} y_{i}\nabla^{2}f_{i}(x^{(l)}) \\ 
&=& \sum_{i=1}^{k} \left(m_l f_{i}(x^{(l)}) -y_{i}\right)\nabla^{2}f_{i}(x^{(l)})\\
&& + m_l J(x^{(l)})^{T}J(x^{(l)}) \nonumber \\[2mm]
&=& m_l J(x^{(l)})^{T}J(x^{(l)})+ W_{\alpha}(x^{(l)}) \in \mathbb{R}^{n \times n}, \nonumber
\end{eqnarray}
\noindent where 

\begin{equation}\label{eq:Walpha_igd}
\begin{split}
W_{\alpha}(x^{(l)}) &= \sum_{i=1}^{k} \underbrace{\left(m_l f_{i}(x^{(l)}) -y_{i}\right)}_{:= \alpha_i^{(l)}} \nabla^{2}f_{i}(x^{(l)})\\[2mm] 
& =  \sum_{i=1}^{k} \alpha_{i}^{(l)} \nabla^{2}f_{i}(x^{(l)}).
\end{split}
\end{equation}

\noindent Thus, the Hessian matrix of $IGD^{2}_{2}$ is given by 

\begin{equation}\label{eq:higd2}
\nabla^{2} IGD^{2}_{2}(X) = \frac{2}{M} \mbox{diag} \left(\mathcal{D}g(x^{(1)}),\ldots, \mathcal{D}g(x^{(\mu)})\right) \in \mathbb{R}^{n\cdot \mu \times n \cdot \mu},
\end{equation}

\noindent which is a block diagonal matrix. \\

As mentioned above, we focus here on the special case $p=2$. The above derivatives, however, can be generalized for $p>1$; for more details, we refer to \cite{uribe:20}.

\section{The $\Delta_p$-Newton Step for Constrained MOPs}

\subsection{The $GD$-Newton Step}

\paragraph{Handling equalities}
See main text. 

\paragraph{Handling inequalities}

For the treatment of the equality constraints, consider that MOP (\ref{eq:mop_suppl})  is subject to 

\begin{equation}\label{eq:ineq}
  g(x) \leq 0. 
\end{equation}

Hereby, let $g:\mathbb{R}^n\to\mathbb{R}^m$, $g(x) = (g_1(x),\ldots, g_m(x))^\top$, 
where $g_i:\mathbb{R}^n \to \mathbb{R}$, $i=1,\ldots, m$, denotes the $i$-th inequality constraint. 
Given a set/population  $X = \{x^{(1)},\ldots, x^{(\mu)}\}\subset\mathbb{R}^n$, the feasibility of 
the associated point  

\begin{equation}
\X = (x^{(1)}_1,\ldots, x^{(1)}_n, x^{(2)}_1,\ldots, x^{(2)}_n, \ldots, x^{(\mu)}_1,\ldots, x^{(\mu)}_n)
\in\mathbb{R}^{\mu n}
\end{equation}

is given if 

\begin{equation}
 g_i(x^{(j)}) \leq 0,\qquad  i = 1,\ldots, m, \; j = 1,\ldots, \mu. 
\end{equation}

To define the population-based inequality function, let 

\begin{equation}
\begin{split}
 g_{i,j} &:\mathbb{R}^{\mu n} \to \mathbb{R}\\
 g_{i,j}(\X) &= g_i(x^{(j)})
\end{split}
\end{equation}

for $i\in\{1,\ldots,m\}$ and $j\in\{1,\ldots, \mu\}$. 
Further, define $\bar{g}:\mathbb{R}^{\mu n}\to\mathbb{R}^{m n}$  by

\begin{equation}
\bar{g}(X) = 
   \begin{pmatrix} g_{1,1}(\X)\\h_{2,1}(\X)\\\vdots\\h_{m,1}(\X)\\h_{1,2}(\X)\\h_{2,2}(\X)\\\vdots\\h_{m,2}(\X)\\
                   \vdots\\h_{m,n}(\X)
   \end{pmatrix}
   =:
   \begin{pmatrix} \bar{g}_1(\X)\\\bar{g}_2(\X)\\\vdots\\\bar{g}_m(\X)\\\bar{g}_{m+1}(\X)\\\bar{g}_{m+2}(\X)\\
                   \vdots\\\bar{g}_{2m}(\X)\\\vdots \\ \bar{g}_{m n}(\X). 
   \end{pmatrix}
\end{equation}

Having defined $\bar{g}$, we are now in the position to explain the simple inequality constraint handling we have 
used in our computations: we say that an inequality constraint is nearly active at $\X$ if 

\begin{equation}
\bar{g}_l(\X) > -\text{tol}, 
\end{equation}

where $\text{tol} > 0$ is a certain tolerance value (hereby, we assume that the value 
$\bar{g}_l(\X)$ is not too large. Recall that we obtain the initial point
$\X_0$ from the run of a MOEA). 
If, in addition, a line search using the search direction $\Delta \X$ will increase the value of $\bar{g}_l$, indicated by 

\begin{equation}
\nabla \bar{g}_l(\X)^\top \Delta \X \geq 0, 
\end{equation}

we add  this constraint to the set of equalities

\begin{equation}
\bar{g}_l(X) = 0. 
\end{equation}

All other inequalities will be disregarded at $\X$, and the Newton step will be performed as for the equality-constrained case. 

\subsection{The $IGD$-Newton Step}

\paragraph{Handling equalities}
Analog to the $GD$ method, the root finding problem for the indicator $IGD_2^2$ reads as
 
\begin{equation}\label{eq:R_IGD=0}
\begin{split}
R\colon& \mathbb{R}^{n(\mu + p)} \to \mathbb{R}^{n(\mu + p)}\\
R(\X,\lambda) &= \begin{pmatrix} \nabla IGD_2^2(\X) + \bar{H}^\top\lambda \\\bar{h}(\X) \end{pmatrix},
\end{split}
\end{equation}
where $\X \in\mathbb{R}^{\mu n}$ and $\lambda \in\mathbb{R}^{\mu p}$. The derivative of $R_{IGD}$ at $(\X,\lambda)^T$ 
is given by 

\begin{equation}
 DR_{IGD}(\X,\dual) = \begin{pmatrix} \nabla^2 IGD_2^2(\X) + S & \bar{H}^\top\\ \bar{H} & 0 \end{pmatrix}, 
\end{equation}
which is a matrix in $\mathbb{R}^{\mu(n+p)\times \mu(n+p)}$.
Also, in this case, we have omitted $S$ so that the Newton step is obtained by solving 

\begin{equation}\label{eq:IGD_newton-step_full}
\begin{split}
  \begin{pmatrix} \nabla^2 IGD_2^2(\X_k) & \bar{H}^\top \\ \bar{H} & 0
  \end{pmatrix}
 \begin{pmatrix} \X_{k+1} - \X_k \\ \dual_{k+1} - \dual_k
  \end{pmatrix}\\[2mm]
 = - 
 \begin{pmatrix} \nabla IGD_2^2(\X_k) + \bar{H}^\top \dual_k \\ \bar{h}(\X_k)
  \end{pmatrix}. 
  \end{split}
\end{equation}

Since also $\nabla^2 IGD_2^2(\X_k)$ has a block structure, we can also, in this case, reduce the
computational complexity significantly: instead of solving (\ref{eq:IGD_newton-step_full}), one
can compute the tuples $(\Delta x_k^{(i)},\Delta \lambda_k^{(i)})$, $i=1,\ldots,\mu$, via solving 

\begin{equation}\label{eq:IGD_newton-step_indiv}
\begin{split}
    \begin{pmatrix} \frac{2}{M} Dg_{IGD}(x_k^{(i)}) & H(x_k^{(i)})^\top \\ H(x_k^{(i)}) & 0
      \end{pmatrix} 
    \begin{pmatrix} \Delta x_k^{(i)} \\ \Delta \lambda_k^{(i)}
      \end{pmatrix}\\[2mm]
     = 
     \begin{pmatrix} -\frac{2}{M}\left(J(x_k^{(i)})^\top(m_i F(x_k^{(i)})-\sum\limits_{j\in I_i} z_j)\right) - H(x_k^{(i)})^\top \lambda_k^{(i)}\\ 
      -h(x_k^{(i)})
      \end{pmatrix}
\end{split}
\end{equation}

where 

\begin{equation}
\label{eq:Dg_IGD}
 Dg_{IGD}(x_k^{(i)}) = m_i J(x_k^{(i)})^\top J(x_k^{(i)}) + \sum_{l=1}^k \alpha_l^{(i)} \nabla^2 f_l(x_k^{(i)})\in\mathbb{R}^{n\times n}. 
\end{equation}

\paragraph{Vanishing sub-gradients}
The following discussion shows that if the set $\X$ contains an element $x^{(i)}$ those sub-gradient of $IGD_2^2$ vanishes,
then the Newton method will iterate $x^{(i)}$ toward a feasible solution. Once feasible, however, the iterates will remain on 
this point and, in particular, not help to improve the $IGD_2^2$ value of the set. \\
Assume that we are given an index $i\in \{1,\ldots, \mu\}$ with 

\begin{equation}
\frac{\partial IGD_2^2(\X)}{\partial x^{(i)}} = 0,
\end{equation}

then (\ref{eq:IGD_newton-step_indiv}) reduces to 

\begin{equation}
\label{eq:IGD_newton-step_indiv_IGD0}
\begin{pmatrix} 0 & H(x^{(i)})^\top \\ H(x^{(i)}) & 0
  \end{pmatrix} 
\begin{pmatrix} \Delta x_k^{(i)} \\ \Delta \lambda_k^{(i)}
  \end{pmatrix}
 = -
 \begin{pmatrix}  H(x^{(i)})^\top \lambda_k^{(i)}\\ 
  h^{(i)}(\X_k)
  \end{pmatrix},
\end{equation}

which is equivalent to 

\begin{equation}
\label{eq:IGD0}
\begin{split}
H(x^{(i)}) \Delta x_k^{(i)}            &= - h^{(i)}(x^{(i)})\\
H(x^{(i)})^\top \Delta \lambda_k^{(i)} &= - H(x^{(i)})^\top \lambda_k^{(i)}.
\end{split}
\end{equation}

Only the first equation in (\ref{eq:IGD0}) has an impact on $x^{(i)}$. We can assume $p<n$ (less equality constraints than decision variables), and hence, this equation is underdetermined. Using $\Delta x_k^{(i)} := - H(x^{(i)})^+ h^{(i)}(x^{(i)})$, where $H(x^{(i)})^+$ denotes the pseudo-inverse of $H(x^{(i)})$, we obtain linear convergence toward a point $x_*^{(i)}$ with $h(x_*^{(i)}) = 0$~\cite{chen1994newton,yamamoto2000historical}. If $x^{(i)}$ is already feasible, then we obtain $\Delta x_k^{(i)}=0$. 

\paragraph{Handling inequalities}
Inequalities are handled as for the $GD$-Newton step. 

\subsection{The $\Delta_p$-Newton Step for General Reference Sets}
The $\Delta_p$-Newton step follows directly from the definition of $\Delta_p$ and the Newton steps
presented above. If the generational distance of $F(\X_l)$ and $Z$ is larger than the respective
value for the inverted generational distance, then the next iterate  $(\X_{l+1},\lambda_{l+1})$ is computed using the 
$GD$-Newton step (\ref{eq:newton-step_indiv}) (of the main text). Else, the $IGD$-Newton step (\ref{eq:IGD_newton-step_indiv})
is used. Algorithm \ref{alg:Deltap_step} shows the pseudo code of this Newton step for general reference
sets $Z$. 

\begin{algorithm}
\caption{$\Delta_p$-Newton Step}
\begin{algorithmic}[1]
\renewcommand{\algorithmicrequire}{\textbf{Input:} current iterate $(\X_l,\lambda_l)$, reference set $Z$}
\renewcommand{\algorithmicensure}{\textbf{Output:} next iterate $(\X_{l+1},\lambda_{l+1})$}
\Require 
\Ensure 
\State Compute $GD_2^2(F(\X_l),Z)$ and $IGD_2^2(F(\X_l),Z)$
\If {$GD_2^2(F(\X_l),Z) < IGD_2^2(F(\X_l),Z)$}
  \State Compute $(\X_{l+1},\lambda_{l+1})$ using (\ref{eq:newton-step_indiv})(main text)
\Else  
  \State Compute $(\X_{l+1},\lambda_{l+1})$ using (\ref{eq:IGD_newton-step_indiv})
\EndIf
\State \Return{ $(\X_{l+1},\lambda_{l+1})$}
\end{algorithmic}
\label{alg:Deltap_step}
\end{algorithm}

\section{The $\Delta_p$-Newton Method within MOEAs}

\subsection{Generating $Z$}

The process of generating the reference set $Z$ involves seven steps:
\begin{enumerate}
    \item[1.] Component detection
    \item[2a.] Filling $(k=2)$
    \item[2b.] Filling $(k > 2)$
    \item[3.] Handling problematic instances
    \item[4.] Generation of the  unshifted reference set $T$
    \item[5.] Computation of the shifting direction $\eta$
    \item[6.] Shift $T$
    \item[7.] Matching
\end{enumerate}
which will be described in detail in the following subsections.

\subsubsection{Component Detection}
Since the PF might be disconnected, we first apply a component detection on the
merged set $P$. 
We used DBSCAN in objective space for this purpose because of three reasons: 
(i) the number of components is not required a priori, (ii) the method detects outliers and 
(iii) we have experienced that a density-based approach works better than a distance-based one 
(e.g., $k$-means) for component detection.
DBSCAN has two parameters: $\var{minpts}$ and $r$. 
To have the component detection ``parameter-free", we compute a small grid-search to find the best values of $\var{minpts}$ and $r$ according to the \textit{weakest link} function defined in~\cite{ClusterQuality}.
The values for the parameters used are $\var{minpts} \in \{2,3\}$ and $r\in\{0.10\bar d,0.11\bar d,\dots,0.15\bar d\}$ for bi-objective problems and $\var{minpts} \in \{3,4\}$ and $r\in\{0.19\bar d,0.20\bar d,\dots,0.23 \bar d\}$ otherwise, where $\bar d$ is the average pairwise distance between all the points. A summary of the component detection process can be seen in Algorithm~\ref{alg:ComponentDetection}. In the following, we describe
the remaining steps for {\em one} connected component. If there are more components, the procedures must be repeated analogously for each $C^i$ found by Algorithm \ref{alg:ComponentDetection}. 

\begin{algorithm}
\caption{Component Detection}
\begin{algorithmic}[1]
\renewcommand{\algorithmicrequire}{\textbf{Input:}}
\renewcommand{\algorithmicensure}{\textbf{Output:}}
\Require merged population $P$, number of objectives $k$ 
\Ensure Number of clusters $\var{nc}$, clusters $ C=\{C^1,\dots,C^{\var{nc}}\}$.
\State Set $\bar{d} \gets 2\sum_{p_i\neq p_j} |p_i-p_j| / (|P'|(|P'|-1))$ 
\If{$k=2$}
\State Set $\var{r\_range}\gets \{0.1\bar{d},0.11\bar{d},\dots,0.16\bar{d}$\} 
\State Set $\var{minpts\_range}\gets \{2,3$\}
\Else
\State Set $\var{r\_range}\gets \{0.19\bar{d},0.20\bar{d},\dots,0.23\bar{d}$\}
\State Set $\var{minpts\_range}\gets \{3,4$\}
\EndIf
\State Set $\var{wlmin}\gets\infty$
\For {$\var{minpts} \textbf{ in } \var{minpts\_range}$}
\For {$r \textbf{ in } \var{r\_range}$}
\State $C_t,\var{nc}_t \gets$ DBSCAN($P',r,\var{minpts}$)
\State $\var{wl} \gets $ WeakestLink($C_t$)
\If {$\var{wl} \leq \var{wlmin}$}
\State $C \gets C_t$
\State $\var{nc} \gets \var{nc}_t$
\State $\var{wlmin} \gets \var{wl}$
\EndIf
\EndFor
\EndFor
\\ \Return{\{$C=\{C^1,\dots,C^{\var{nc}}\},\var{nc}$\}}
\end{algorithmic}
\label{alg:ComponentDetection}
\end{algorithm}

\subsubsection{Filling}
The final populations of an evolutionary algorithm do not have to be uniformly distributed along
the PF (and the same may hold for the merged set $P$). However, if we fill the gaps and select $\mu$ points out of 
the filled set, we can achieve a more uniform set and, in turn, better IGD approximations when 
making a selection out of these filled sets. The next task is, hence, to compute $N_f$ solutions 
that are ideally uniformly distributed along the computed part of the solution set. 
This process is done differently for $k=2$ and $k\geq3$ objectives.
\setcounter{subsubsection}{1}
\subsubsection{Filling ($k=2)$} For $k=2$ we sort the points of $P = \{p_1,\dots,p_\ell\}$ in increasing order of objective $f_1$. Then, we consider the piece-wise linear curve resulting from $p_1$ and $p_2$, $p_2$ and $p_3$, and
    so on. The total length of this curve is given by $|L|=\sum_{i=1}^{\ell-1} |L_i|$, where $ |L_i| =\|p_i-p_{i+1}\|_2$. 
    To realize the filling, we arrange the $N_f$ desired points along the curve $L$ such that the first point is $p_1$ and the following points are distributed equidistantly along  $L$. This is accomplished by setting each point a distance of $\delta_\ell = |L|/(N_f-1)$ apart from each other along $L$. See Algorithm \ref{alg:fillingk=2} for details.
\setcounter{subsubsection}{1}
\subsubsection{Filling ($k > 2)$} The filling for $k > 2$ has several intermediate steps that we need to describe first. See Algorithm \ref{alg:fillingk>3} for a general procedure outline. 
    First, to better represent $P$ (particularly for
    the filling), we triangulate this set in $k-1$ dimensional space. This is done since we know the PF for continuous MOPs is an at most $k-1$-dimensional object.  In order to do this, we compute a ``normal vector'' $\eta$ to $P$ using 
    equation (\ref{eq:eta}), and then we project it into a the $k-1$ hyperplane normal 
    to $\eta$, obtaining the projected set $P_{k-1}$. After this, we compute the Delaunay triangulation \cite{Delaunay_1934aa} of $P_{k-1}$, which gives us a triangulation $DT$ that can be used in the original $k$-dimensional space. Finally, each triangle (or simplex for $k>3$) $\Theta_i\in DT$ is filled at random with the number of points proportional to its area (respectively volume for $k>3$) so that we end up with the filled set $Y$ of size $N_f$. \\
    We will describe each step in more detail in the following paragraphs:
    
\paragraph{Compute ``normal vector'' $(\eta)$} using equation (\ref{eq:eta}). We
          use the same vector for the shifting direction, which we will describe below.
\paragraph{$k-1$ Projection $(P_{k-1})$} We can use $\eta$ as the first axis of a new coordinate system $(\eta,v_1,\dots,v_{k-1})$. By doing so, the orthonormal vectors $v_1,\dots,v_{k-1}$ become the basis of an orthogonal hyperplane to $\eta$, and the projection of $P$ onto this hyperplane ($P_{k-1}$) can be achieved by first changing $P$ into the coordinate system obtaining $P = \beta\eta + \beta_1v_1 + \dots + \beta_{k-1}v_{k-1} $, and then removing the first coordinate, i.e., $P_{k-1} = \beta_1v_1 + \dots + \beta_{k-1}v_{k-1}$. The computation of the vectors 
        $v_i$ will be described below (see Step 5 \textit{Computation of the Shifting Direction $\eta$)}.
\paragraph{Delaunay Triangulation $(DT)$} Compute the Delaunay triangulation of $P_{k-1}$. This returns $DT$, a list of the indices of $P_{k-1}$ that form the triangles (simplices for $k>3$). This list $DT$ is the triangulation used for the $k$-dimensional set $P$, which is possible because $DT$ is a list of indices, i.e., it is independent of the dimension. 

\paragraph{Triangle Filling $(Y)$} For each triangle $\Theta_i \in DT$ of area $a_i$, $\lceil a_iF / A\rceil$ points are generated uniformly. At random inside of the triangle $\Theta_i$ following the procedure of \cite{UniformSamplingSimplex} to obtain the unshifted reference set $Y$.

\begin{algorithm}
\caption{Filling ($k=2$ Objectives)}
\begin{algorithmic}[1]
\renewcommand{\algorithmicrequire}{\textbf{Input:}}
\renewcommand{\algorithmicensure}{\textbf{Output:}}
\Require Population $P'=\{p_1',\ldots,p_\ell' \} $, filling size $N_f$
\Ensure Filled set $Y=\{y_1,\dots,y_{N_f}\}$

\State $X = \{x_1,\dots,x_\ell\} \gets$ sort $P'$ according to its first objective $f_1$ 
\State $L_i \gets \|x_i-x_{i+1}\|_2 \quad \forall i=1,\dots,\ell-1$
\State $L\gets \sum_{i=1}^{\ell-1} L_i$
\State $\delta_\ell \gets L/(N_f-1)$
\State $\var{dist\_left} \gets (0,0,\dots,0)\in \mathbb{R}^\ell$
\For{$i=1:\ell-1$} \Comment{\footnotesize compute number of points per segment} \normalsize
    \State $\var{ratio} \gets (|L_i|+\var{dist\_left}(i))/\delta_\ell$ 
    \State $\var{points\_per\_segment}(i) \gets \lfloor \var{ratio} \rfloor$
    \State $\var{dist\_left}(i+1) \gets (\var{ratio}-\lfloor \var{ratio} \rfloor)\delta_\ell$
\EndFor
\State $\var{count} = 1$
\For{$i = 1:\ell-1$} \Comment{\footnotesize for each line segment} \normalsize
    \If{$\var{points\_per\_segment}(i)>0$} \Comment{\footnotesize check if  any point lands in segment $L_i$} \normalsize
        \State $\nu_i := (x_{i+1}-x_{i})/L_i$
        \State $y_{\var{count}} = x_i + (\delta_\ell-\var{dist\_left}(i)) \nu_i$
        \State $\var{count} \gets \var{count} + 1$
        \For{$j=2:\var{points\_per\_segment}(i)$} \Comment{\footnotesize if $L_i$ has more than one point} \normalsize
            \State $y_{\var{count}} \gets y_{\var{count}-1} + \delta_\ell \nu_i$
            \State $\var{count} \gets \var{count} + 1$
        \EndFor
    \EndIf
\EndFor \\
\Return{$Y$}
\end{algorithmic}
\label{alg:fillingk=2}
\end{algorithm}

\begin{algorithm}
\caption{Filling ($k\geq 3$ Objectives)}
\begin{algorithmic}[1]
\renewcommand{\algorithmicrequire}{\textbf{Input:}}
\renewcommand{\algorithmicensure}{\textbf{Output:}}
\Require Population $P'=\{p_1',\ldots,p_\ell' \} $, filling size $N_f$
\Ensure Filled set $Y=\{y_1,\dots,y_{N_f}\}$
\State $\eta \leftarrow \var{ComputeNormalVector}(P')$ \Comment{Eq.~\eqref{eq:M-matrix}~\eqref{eq:eta}}
\State $P_{k-1} \leftarrow \text{ project } P' \text{ onto } \eta^\perp$
\State $DT \leftarrow \var{DelaunayTriangulation}(P_{k-1})$
\State $Y \leftarrow \var{TriangleFilling}(DT,F)$
\\ \Return{$Y$}
\end{algorithmic}
\label{alg:fillingk>3}
\end{algorithm}

\subsubsection{Handling Problematic Instances}
There are instances where we get 
less than $\mu$ non-dominated points even after merging the populations. We can summarize these cases as follows:

\begin{itemize}
    \item $|P|<0.1\mu$. In this case, we do not apply the Newton method.
    \item $|P|<0.3\mu$. We have observed that this case appears in two main cases: (i) many of
      the candidate solutions of $P'$ were removed since they have been considered unpromising, and
    (ii) many elements of the considered populations $P_f, P_{f-s}, \ldots, P_{f-(k-1)s}$ are  
      identical. Such scenarios occur, e.g., when dealing with problems where the Pareto front is discrete 
      (e.g., for CF1) or when a reference vector-based MOEA is used as a base algorithm. In both cases, 
       we have observed that the solutions of $P$ are, in many cases, widely spread in objective space. 
       For such scenarios, it was better to relax the parameters of the component detection algorithm 
       to get fewer connected components (many small components will be detected otherwise). Specifically, 
       we have used  for such cases $r\in \{0.5\bar{d},0.6\bar{d}\}$ and $minpts = 2$ for $k=2$, and $r\in \{0.75\bar{d},0.0.77\bar{d},0.0.79\bar{d}\}$ and $minpts = 2$ for $k>2$.
    \item $0.3\mu < |P| < \mu$. We follow the procedure described in the main text but use repeated starting
      points in $\X_0$ as described below (which holds for all problematic instances). 
\end{itemize}

Since we require that $\X_0$ contains $\mu$ elements, for these problematic instances, we can not follow the procedure for computing $\X_0$ as in the main text. Instead, we first set $\X_0 = P$, and then fill the remaining solutions with 
repeated solutions taken from $P$ at random.

\subsubsection{Generation of the Unshifted Reference Set $T$}

Once we have computed the filled set $Y$, we need to select $\mu$ points out of it that are ideally equally distributed along
$Y$. To this end, we use $k$-means with $\mu$ clusters for this purpose due to the relationship
between $k$-means and the optimal $IGD$ subset selection (\hspace{1sp}\cite{uribe:20,Ishibuchi_IGD_clustering}). The obtained $\mu$ centroids form
the reference set $T = \{t_1,\ldots,t_{\mu}\}$.

\subsubsection{Computation of the Shifting Direction $\eta$}
The ideal scenario is certainly to shift each of the target values $t_i$ ``orthogonal'' to the Pareto front toward an
utopian region. Since the front is not known, we compute this shifting direction $\eta$ orthogonal to the convex hull 
defined by the minimal elements of the target values. More precisely, we compute $\eta$ as follows: if 
$\tau=\{t_1,\ldots,t_{\mu}\}$ consists of one connected component  (else perform the following steps on each of the components), choose 
\begin{equation}
    y^{(i)} = t_{j^*_i} ,\quad i=1,\ldots, k,
\end{equation}

where 
\begin{equation}
 j^*_i\in \arg\min_{j=1,\ldots,\mu} t_{j,i},\quad i=1,\ldots, k,
\end{equation}
and $t_{j,i}$ denotes the $i$-th element of $t_j$. Then, set 
\begin{equation}
 M := (y^{(2)} - y^{(1)},y^{(3)} - y^{(1)}, \ldots, y^{(k)} - y^{(1)})\in\mathbb{R}^{k\times(k-1)}. 
\end{equation}
Next, compute a QR-factorization of $M$, i.e., 
\begin{equation} \label{eq:M-matrix}
 M = QR = (q_1,\ldots,q_k)R,
\end{equation}
where $Q\in\mathbb{R}^{k\times k}$ is an orthogonal matrix with column vectors $q_i$, and $R\in\mathbb{R}^{k\times(k-1)}$ 
is a right upper triangular matrix. Then, the vector
\begin{equation}
\label{eq:eta}
\eta = -\sgn(q_{k,1}) \frac{q_k}{\|q_k\|_2}
\end{equation}
is the desired shifting direction. Since $Q$ is orthogonal,
the vectors $v_1 := q_1,\ldots,v_{k-1}:= q_{k-1}$ form an 
orthonormal basis of the hyperplane that is orthogonal to 
$\eta$. That is, these vectors can be used for the construction
of $P_{k-1}$.

\subsubsection{Shift $T$}
The reference set $T$ is shifted following the direction of $\eta$ to a utopian region to obtain $Z$ via $ z_i = t_i + t\eta$. Hereby, $t$ controls how far $T$ is shifted down
(we used $t=0.05$).

\subsubsection{Matching}
 We now have computed two sets of cardinality $\mu$: the initial points $X_0$ for the Newton method and the reference set $Z$. 
Since the latter set has been computed to yield a good representation (in terms of IGD) of the parts of the Pareto 
front that are already detected by the MOEA, it serves as a good target for the Newton method. $X_0$ and $Z$ might not be
matched (in the sense defined in the main text). Therefore, we perform a matching (in the sense of \textit{perfect matching} as in \cite[Chapter~7]{kleinberg_algorithm}, obtained via solving a linear assignment problem (\cite{assignmentProblem}) ) between the two sets to obtain the best possible one-to-one relationship. More precisely, perfect matching is used to determine for each 
$x^{(i)}\in \X_0$ the target $z_{j_i}$ of $Z$ for the Newton method.

\subsection{Complexity Analysis} 

We now present the complexity of each step separately for connected PFs. If the PF is disconnected, the complexity is the same as described below for each component, with the total size being replaced by each component size accordingly.

\begin{enumerate}
    \item[a)] Computing $X_0$. The time complexity is $\mathcal{O}( (\kappa\mu)^2 + \kappa\mu k \tau_1)$ due to the complexity of the non-dominance test plus the $k$-medoids complexity. Here, $\mu$ is the population size, $\kappa$ is the number of populations we merge, $k$ is the number of objectives, and $\tau_1$ is the number of iterations of $k$-medoids (usually small). 
    
    \item[b1)] Generating $Z$ (Component Detection). The time complexity is $\mathcal{O}(\gamma\kappa^2\mu^2)$ due to the size of the grid search ($\gamma$) times the complexities of DBSCAN and the WeakestLink computation. Here, $\gamma$ is the number of parameter combinations of the grid search, which is $\gamma = 14$ for $k=2$ and $\gamma = 10$ for $k\geq 3$.
    
    \item[b2)] Generating $Z$ (Filling). The time complexity depends on the number of objectives:
    \begin{itemize}
        \item For $k=2$ the time complexity is $\mathcal{O}(\ell + kN_f )$ due to sorting and placing the $N_f$ points along the line segments. Here, $\ell$ is the size of the merged population $P$ (without the unpromising solutions).
        \item For $k\geq 3$ the time complexity is $\mathcal{O}(k^3 + \ell k^2 + \ell\log\ell + k\sigma + kN_f)$ due to computing the normal vector $\eta$, the change of coordinates and projection, the Delaunay triangulation and the triangle filling. Here, $\sigma$ represents the size of the Delaunay triangulation, i.e., the number of triangles.
    \end{itemize}
    
    \item[c)] Generating Unshifted Reference Set $T$. The time complexity is $\mathcal{O}( \tau_2 k\mu N_f )$ due to $k$-means. Here, $N_f$ is the number of points of the filled set ($Y$), and $\tau_2$ is the number of iterations of $k$-means.
    
    \item[d)] Computing Shifting Direction. The time complexity is $\mathcal{O}(k\mu+k^3)$ due to the complexity of computing the $k$ minima and of the QR factorization.

    \item[e)] Shifting $T$. Due to matrix addition, the time complexity is $\mathcal{O}(k\mu)$.
    
    \item[f)] Matching. The time complexity is $\mathcal{O}(\mu^2\log\mu)$ due to solving a linear assignment problem to obtain the matching.
    
    \item[g)] $\Delta_p$-Newton Method. The time complexity is $\mathcal{O}(\mu (n+p)^3)$, as discussed above. Here, $n$ is the number of decision variables, and $p$ is the number of equality constraints.
\end{enumerate}

The complexity of the entire method is $\mathcal{O}(\tau_2 k\mu N_f + \mu^2\log\mu + \mu (n+p)^3)$ due to selecting the reference set $T$, the matching and the $\Delta_p$-Newton method  (assuming that $\ell$ is proportional to $\mu$ and $k\ll\mu$).

\subsection{Examples}

\begin{figure} 
    \centering
  \subfloat[$F(P_1)$]{%
       \includegraphics[width=0.5\linewidth]{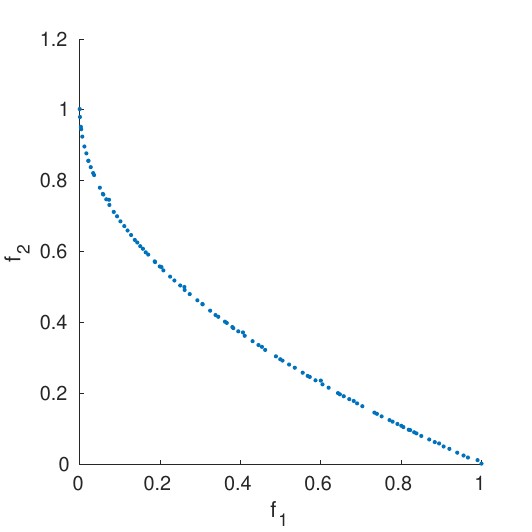}}
    \subfloat[$F(P_2)$]{%
       \includegraphics[width=0.5\linewidth]{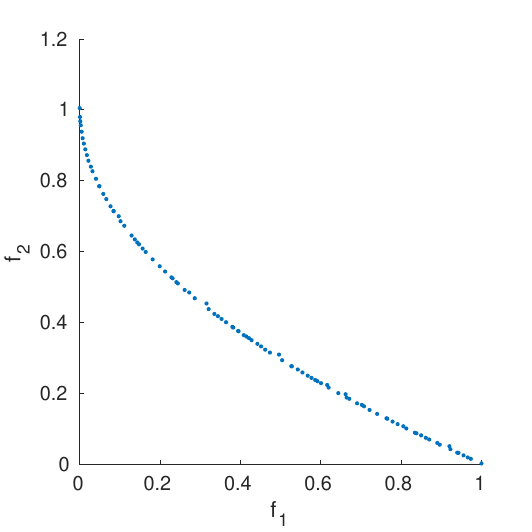}}
  \\
  \subfloat[$Y$ (in blue), T (black dots).]{%
        \includegraphics[width=0.5\linewidth]{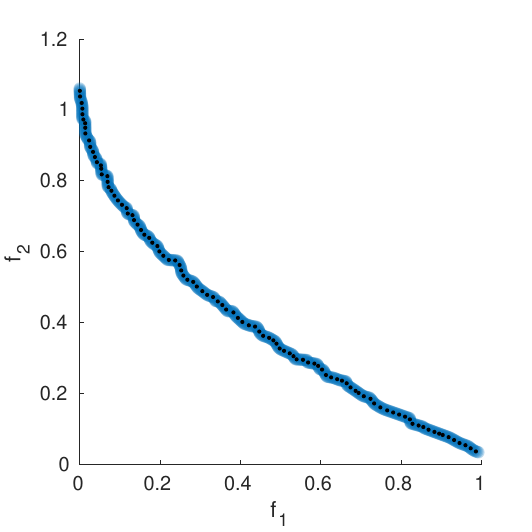}}
  \subfloat[Matching of $\X_0$ (black dots) with $Z$ (purple diamonds)]{%
        \includegraphics[width=0.5\linewidth]{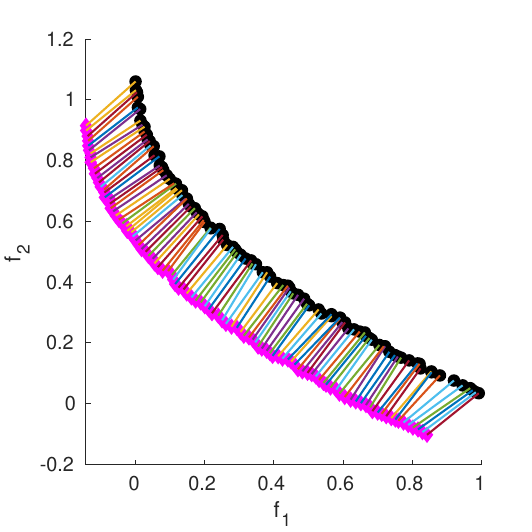}}
  \\
  \subfloat[Newton]{%
        \includegraphics[width=0.5\linewidth]{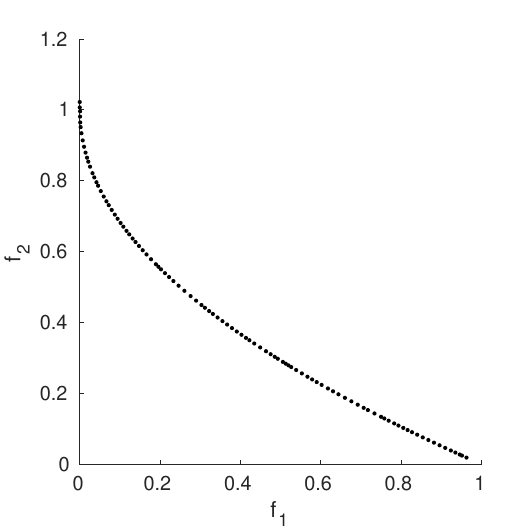}}
  \subfloat[Left axis: IGD, right: $||R(X)||$]{%
        \includegraphics[width=0.53\linewidth, trim=25mm 8mm 15.5mm 20mm, clip]{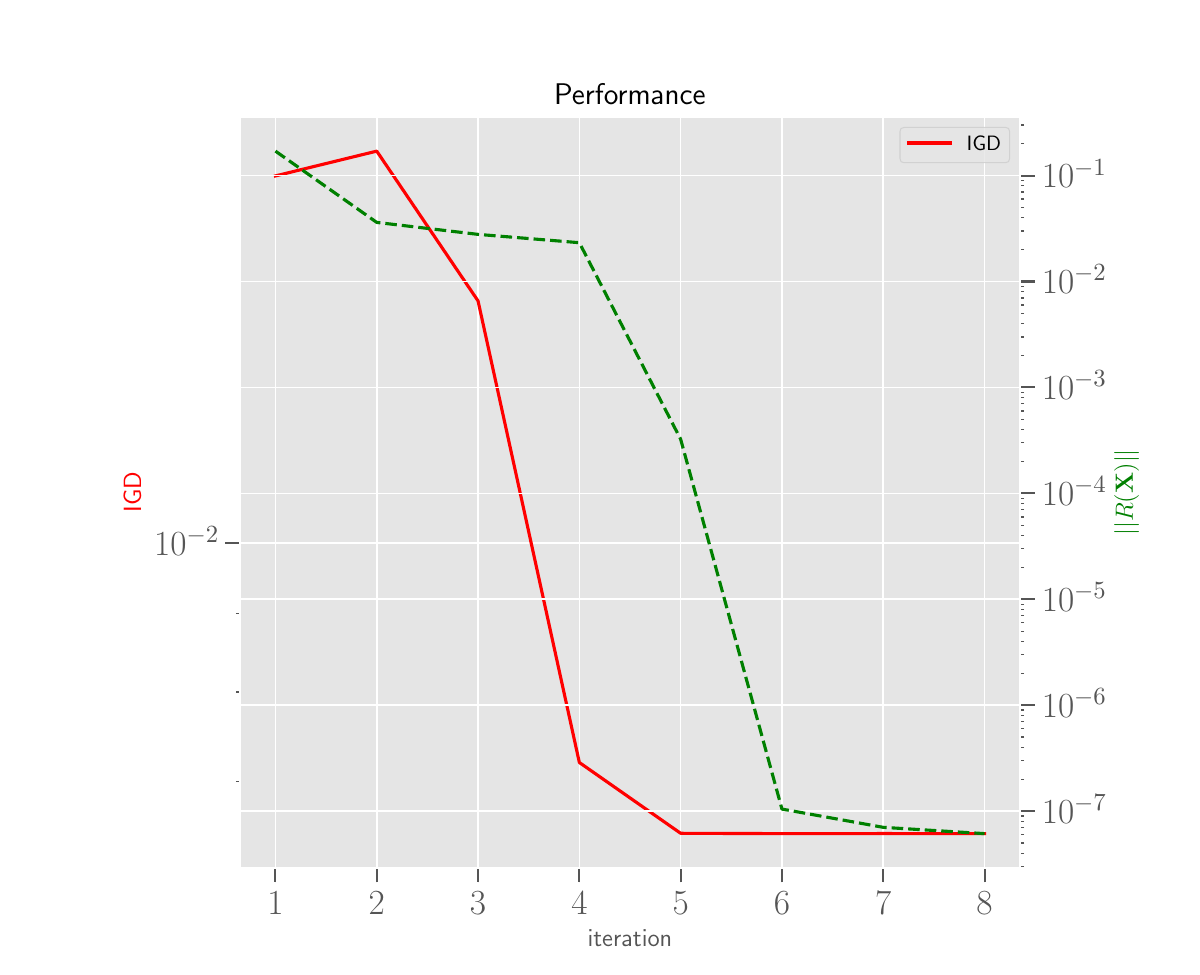}}
  \caption{Application of the $\Delta_p$-Newton method on populations obtained with NSGA-II for ZDT1.}
  \label{fig:ZDT1_EA} 
\end{figure}

\begin{figure} 
    \centering\captionsetup{width=.45\linewidth}
  \subfloat[$F(P_1)$]{%
       \includegraphics[width=0.5\linewidth]{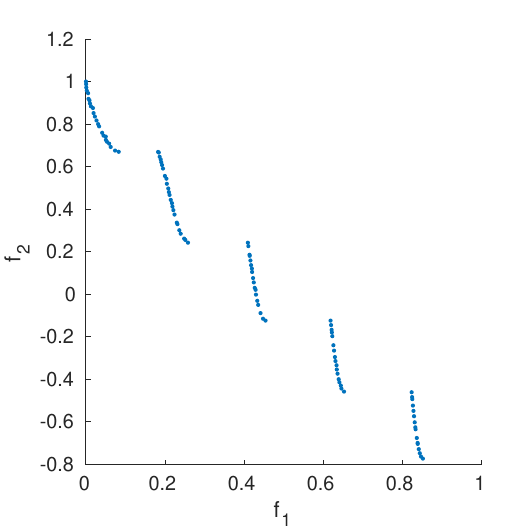}}
    \subfloat[$F(P_2)$]{%
       \includegraphics[width=0.5\linewidth]{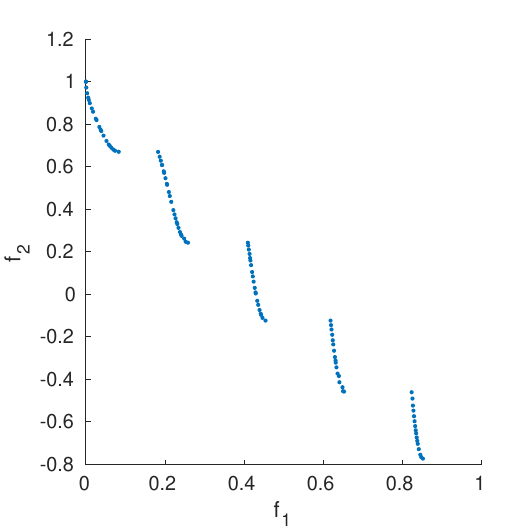}}
  \\
  \subfloat[$Y$ (different colors per connected component), T (black dots).]{%
        \includegraphics[width=0.5\linewidth]{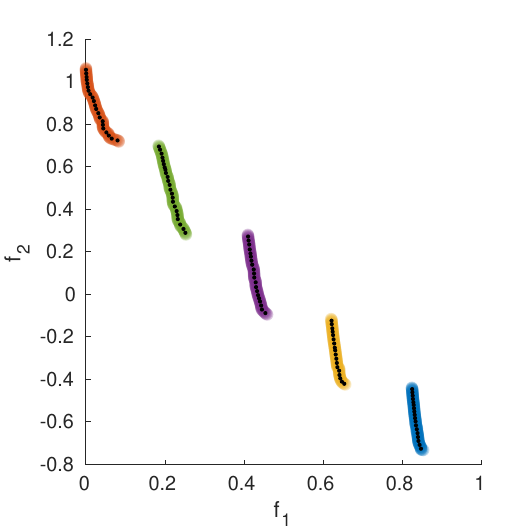}}
  \subfloat[Matching of $\X_0$ (black dots) with $Z$ (purple diamonds)]{%
        \includegraphics[width=0.5\linewidth]{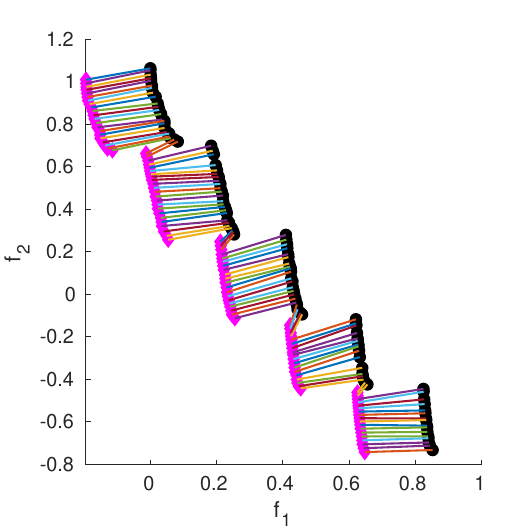}}
  \\
  \subfloat[Newton]{%
        \includegraphics[width=0.5\linewidth]{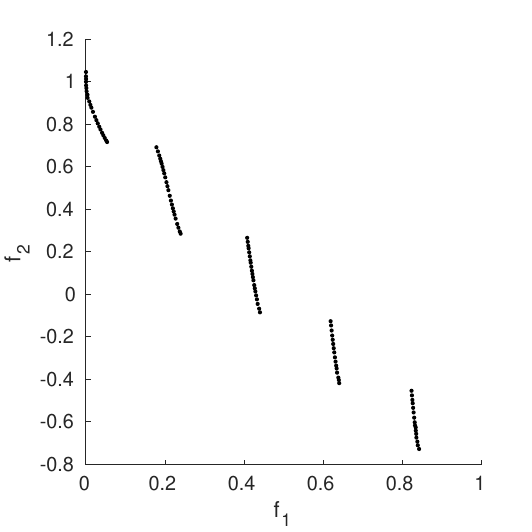}}
  \subfloat[Left axis: IGD, right: $||R(X)||$]{%
        \includegraphics[width=0.57\linewidth, trim=10mm 8mm 15.5mm 20mm, clip]{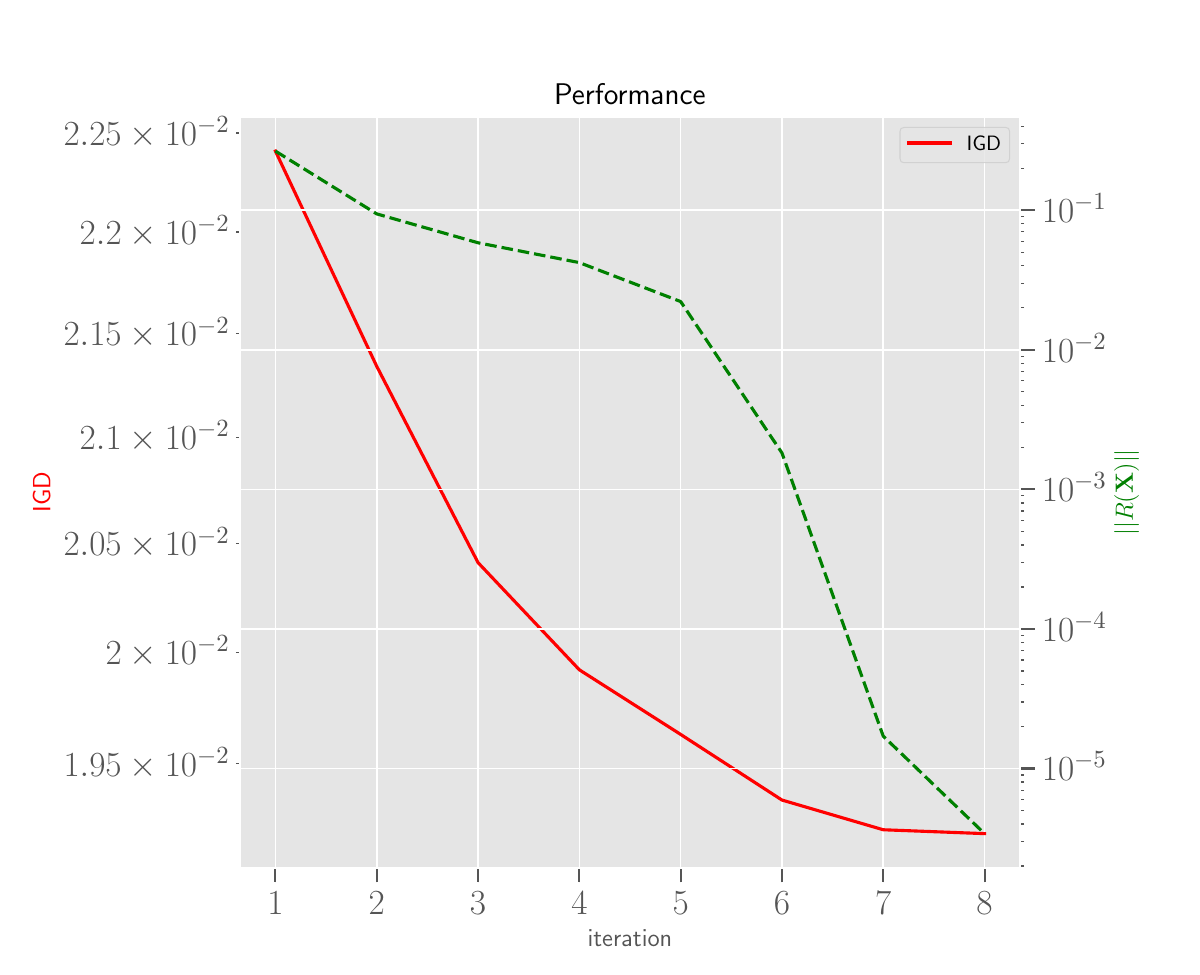}}
  \caption{Application of the $\Delta_p$-Newton method on populations obtained with NSGA-II for ZDT3.}
  \label{fig:ZDT3_EA} 
\end{figure}

\begin{figure} 
    \centering\captionsetup{width=.45\linewidth}
  \subfloat[$F(P_1)$]{%
       \includegraphics[width=0.5\linewidth]{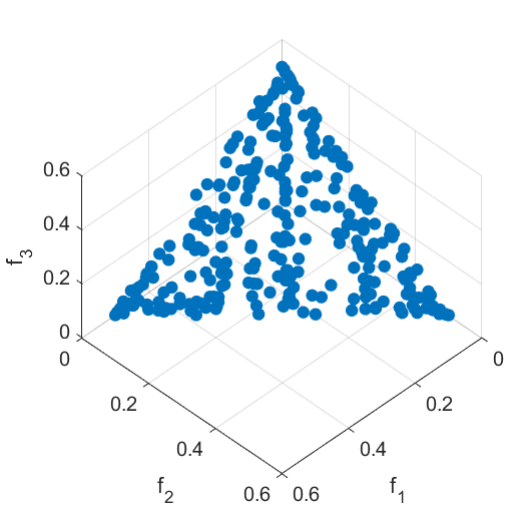}}
    \subfloat[$F(P_2)$]{%
       \includegraphics[width=0.5\linewidth]{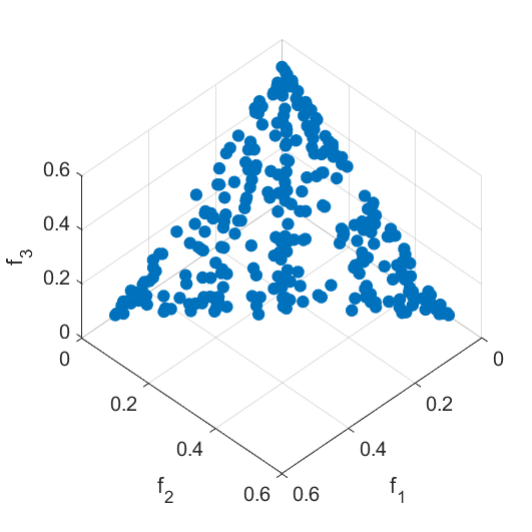}}
  \\
  \subfloat[$Y$ (in blue), T (purple diamonds).]{%
        \includegraphics[width=0.5\linewidth]{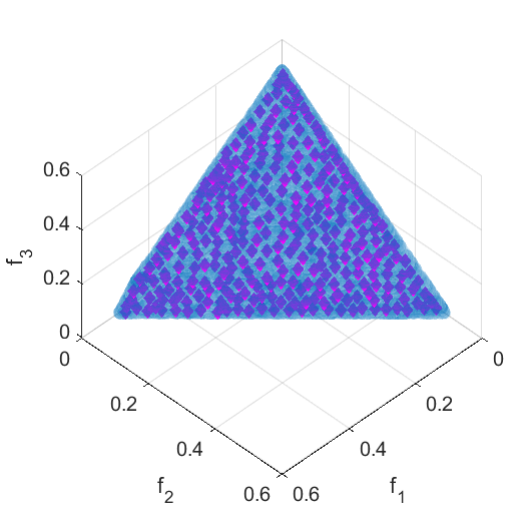}}
  \subfloat[Matching of $\X_0$ (black dots) with $Z$ (purple diamonds)]{%
        \includegraphics[width=0.5\linewidth]{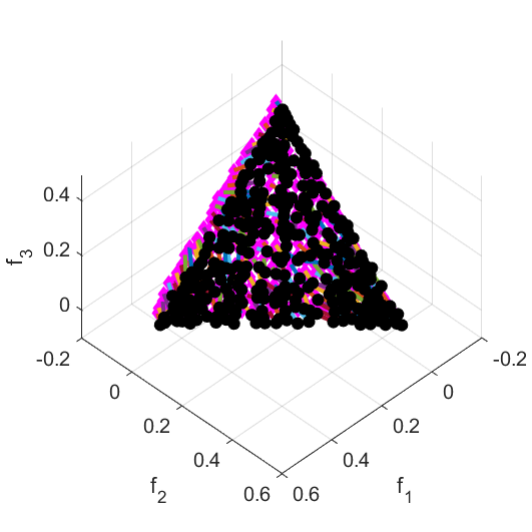}}
  \\
  \subfloat[Newton]{%
        \includegraphics[width=0.5\linewidth]{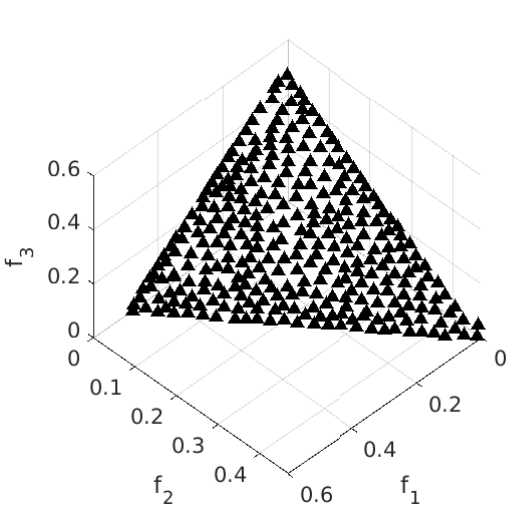}}
  \subfloat[Left axis: IGD, right: $||R(X)||$]{%
        \includegraphics[width=0.55\linewidth, trim=10mm 9mm 15.5mm 20mm, clip]{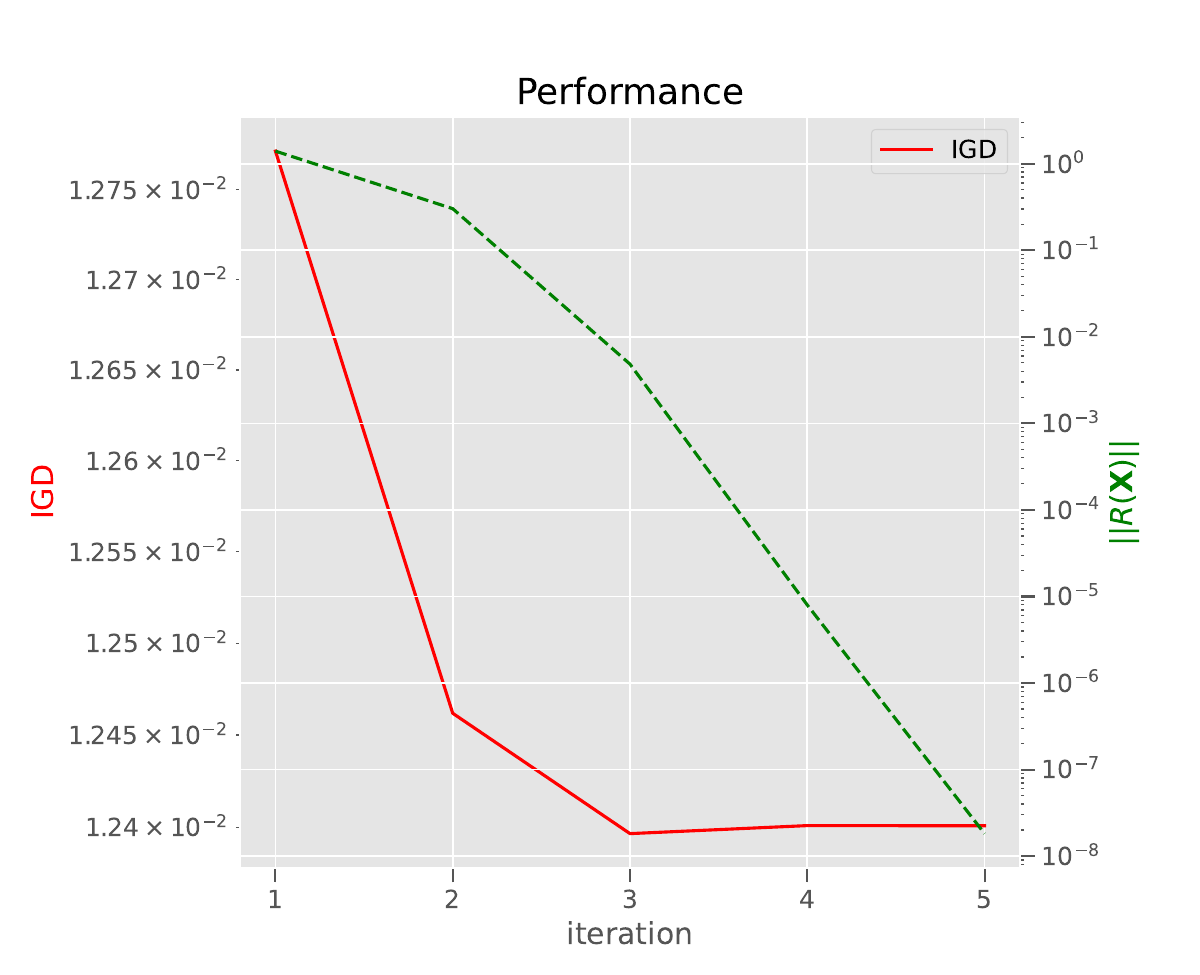}}
  \caption{Application of the $\Delta_p$-Newton method on populations obtained with NSGA-II for DTLZ1.}
  \label{fig:DTLZ1_EA} 
\end{figure}

\begin{figure} 
    \centering\captionsetup{width=.45\linewidth}
  \subfloat[$F(P_1)$]{%
       \includegraphics[width=0.5\linewidth]{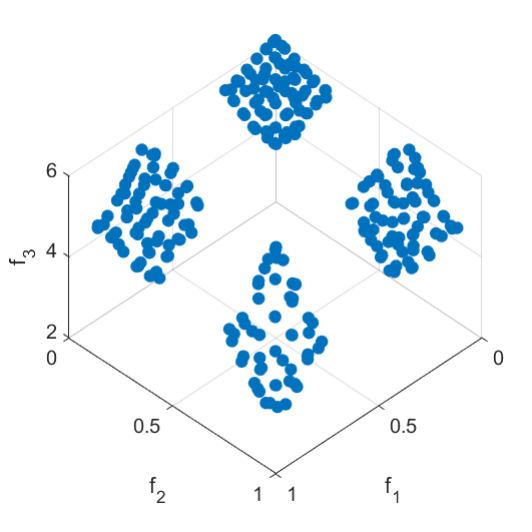}}
    \subfloat[$F(P_2)$]{%
       \includegraphics[width=0.5\linewidth]{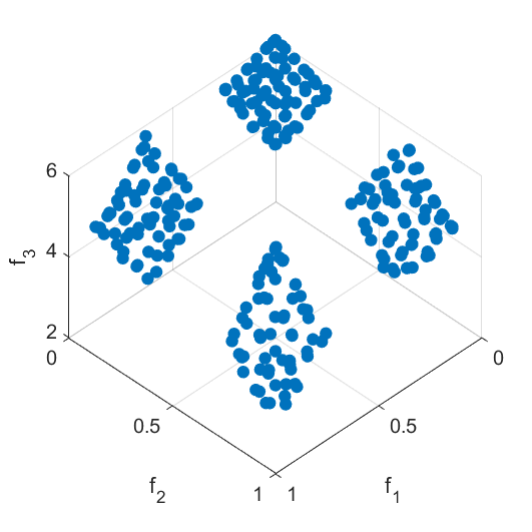}}
  \\
  \subfloat[$Y$ (different colors per connected component), T (purple diamonds).]{%
        \includegraphics[width=0.5\linewidth]{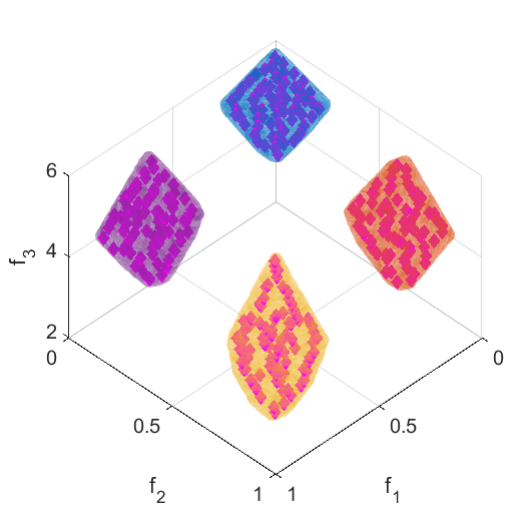}}
  \subfloat[Matching of $\X_0$ (black dots) with $Z$ (purple diamonds)]{%
        \includegraphics[width=0.5\linewidth]{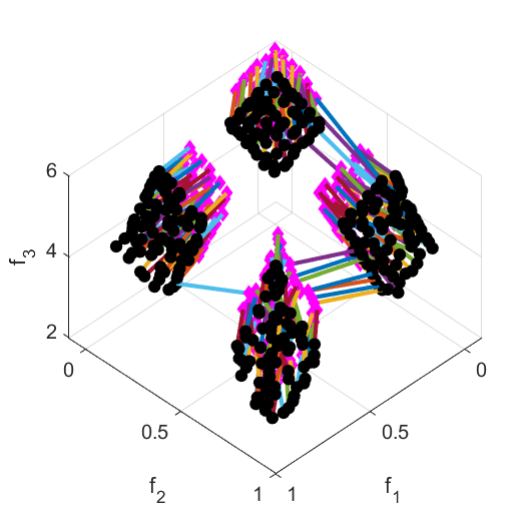}}
  \\
  \subfloat[Newton]{%
        \includegraphics[width=0.5\linewidth]{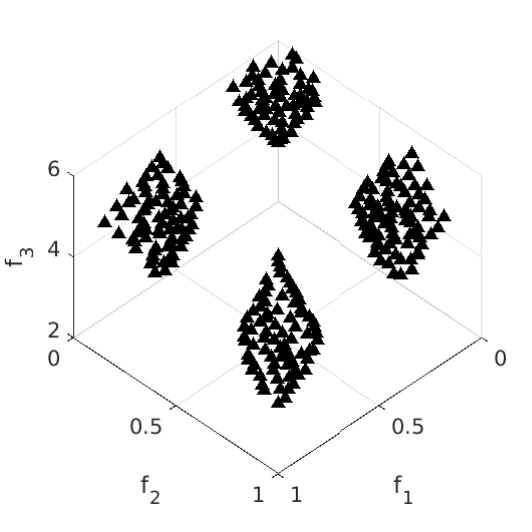}}
  \subfloat[Left axis: IGD, right: $||R(X)||$]{%
        \includegraphics[width=0.545\linewidth, trim=17mm 9mm 17mm 21mm, clip]{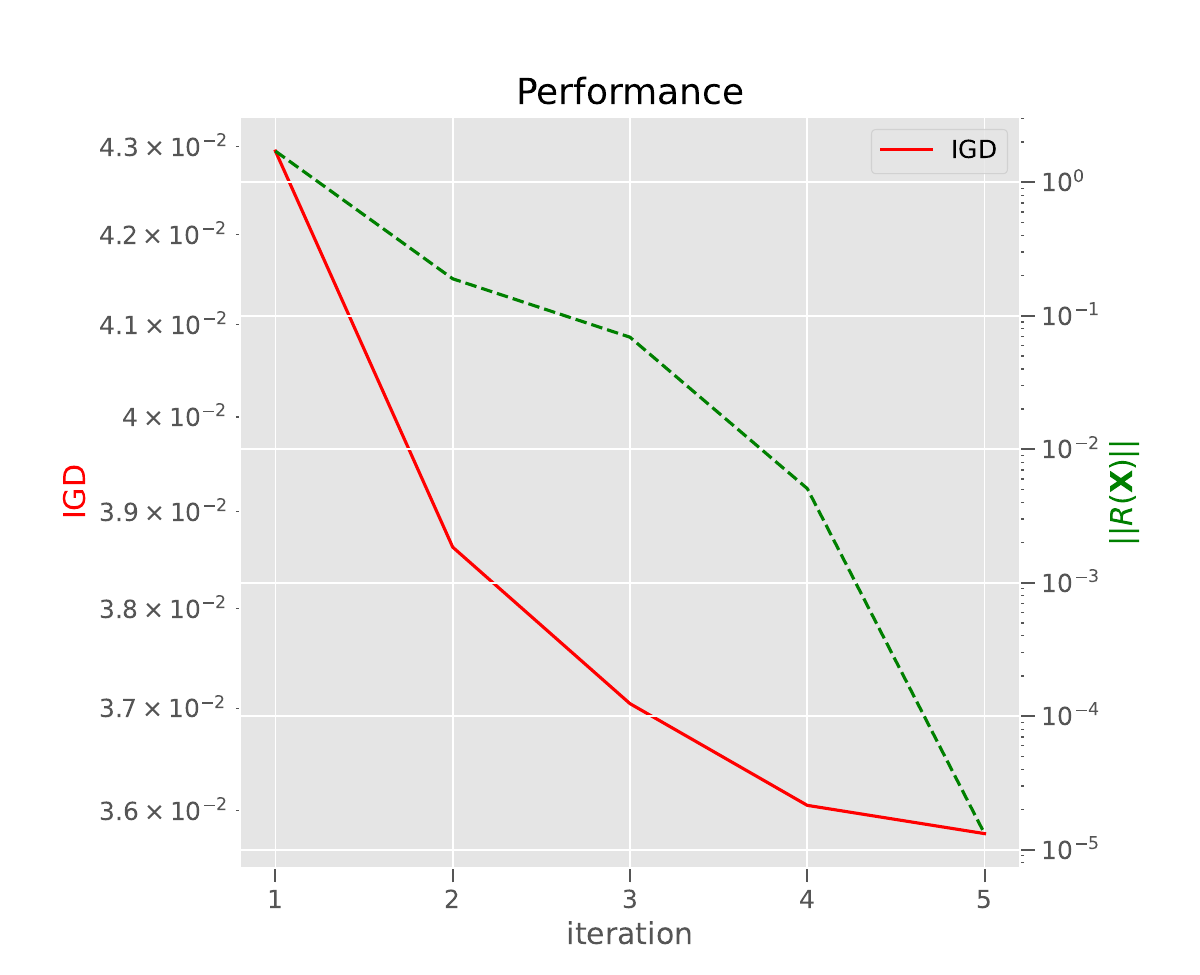}}
  \caption{Application of the $\Delta_p$-Newton method on populations obtained with NSGA-III for DTLZ7.}
  \label{fig:DTLZ7_EA} 
\end{figure}

\begin{figure} 
    \centering\captionsetup{width=.45\linewidth}
  \subfloat[$F(P_1)$]{%
       \includegraphics[width=0.5\linewidth]{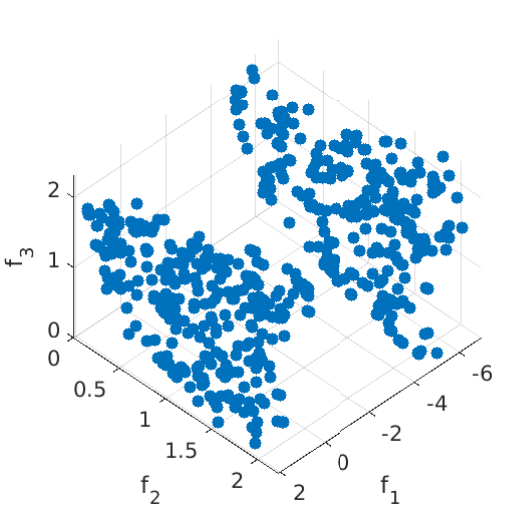}}
    \subfloat[$F(P_2)$]{%
       \includegraphics[width=0.5\linewidth]{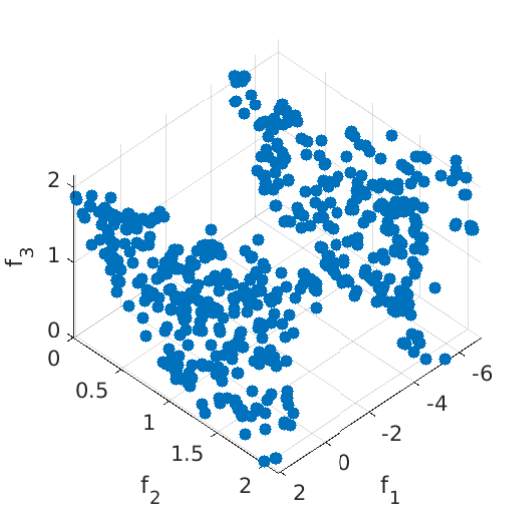}}
  \\
  \subfloat[$Y$ (in blue and red, different colors per connected component), T (purple diamonds).]{%
        \includegraphics[width=0.5\linewidth]{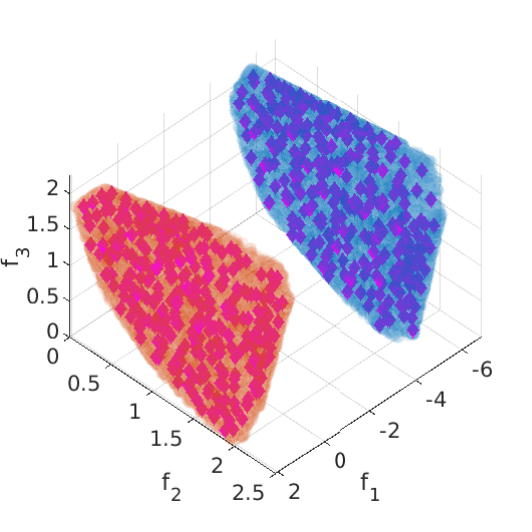}}
  \subfloat[Matching of $\X_0$ (black dots) with $Z$ (purple diamonds)]{%
        \includegraphics[width=0.5\linewidth]{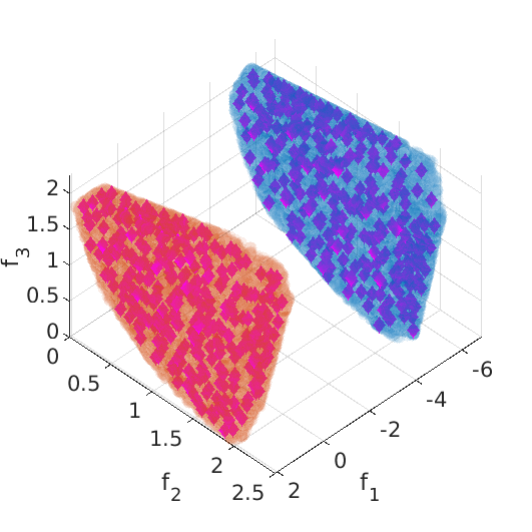}}
  \\
  \subfloat[Newton]{%
        \includegraphics[width=0.5\linewidth]{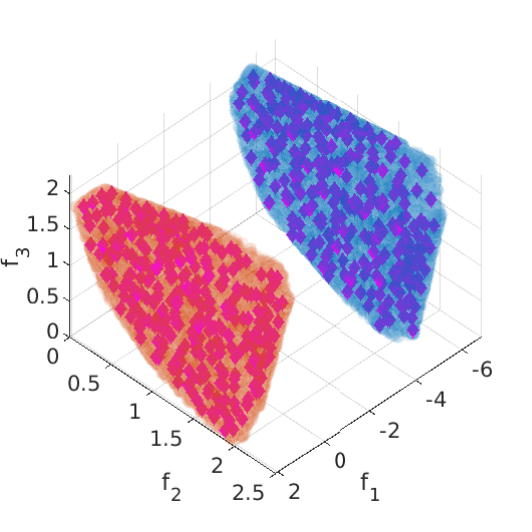}}
  \subfloat[Performance]{%
        \includegraphics[width=0.5\linewidth]{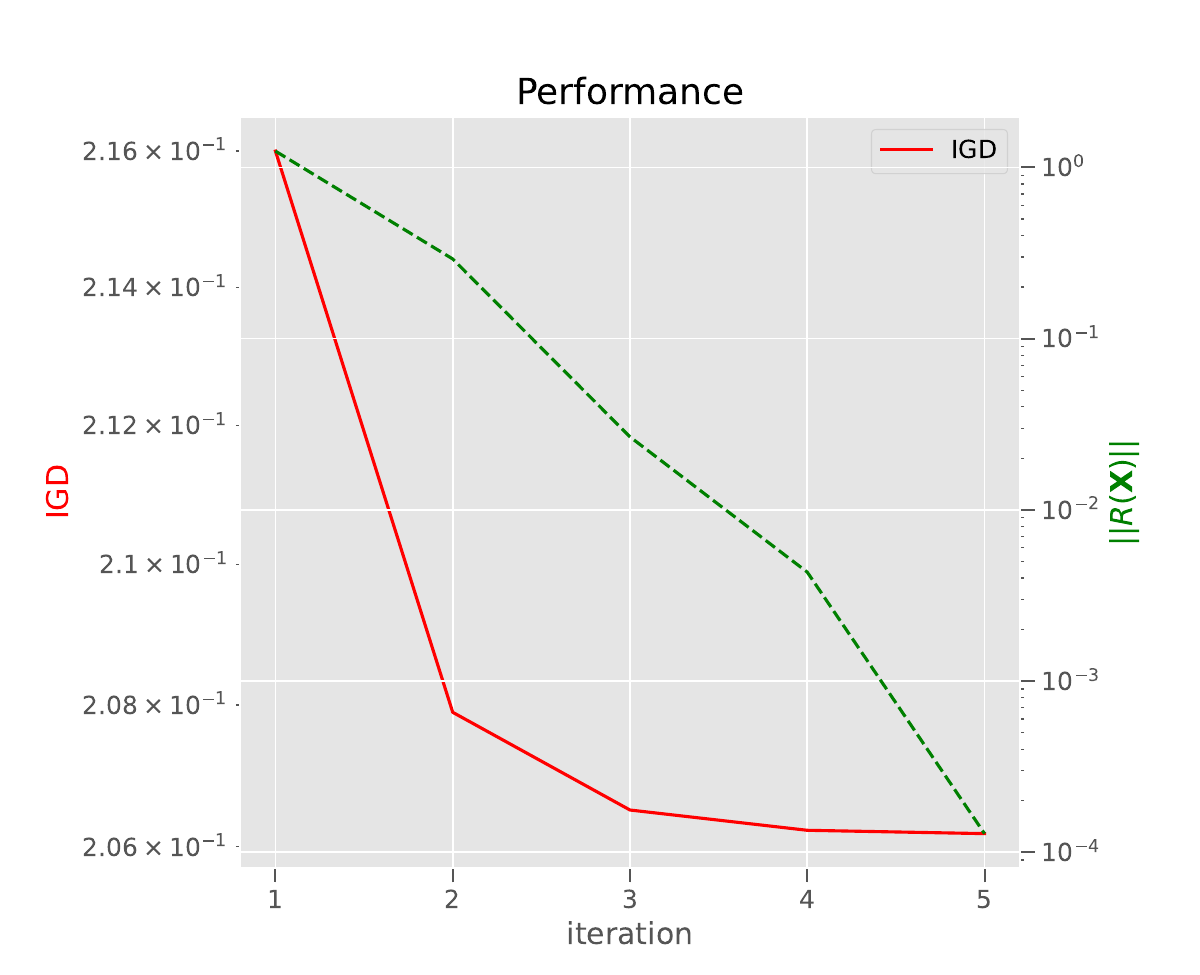}}
  \caption{Application of the $\Delta_p$-Newton method on populations obtained with SMS-EMOA for CONV4-2F. The plots are the projection of objectives $f_1$, $f_2$ and $f_3$. }
  \label{fig:CONV4-2F_EA_123} 
\end{figure}

\begin{figure} 
    \centering\captionsetup{width=.45\linewidth}
  \subfloat[$F(P_1)$]{%
       \includegraphics[width=0.5\linewidth]{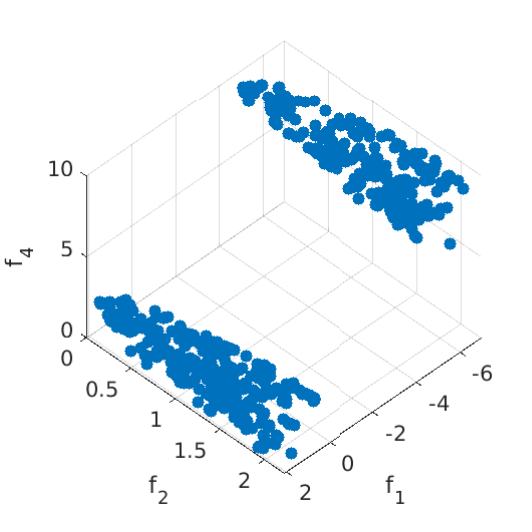}}
    \subfloat[Newton]{%
       \includegraphics[width=0.5\linewidth]{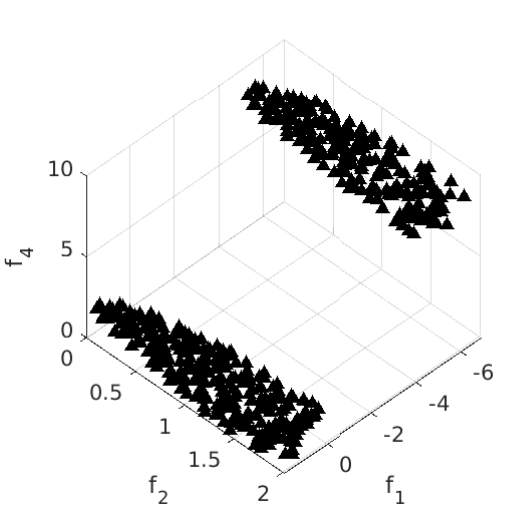}}
  \\
  \subfloat[$F(P_1)$]{%
       \includegraphics[width=0.5\linewidth]{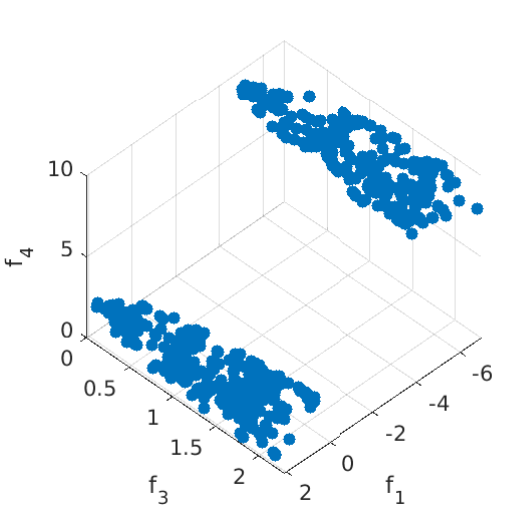}}
    \subfloat[Newton]{%
       \includegraphics[width=0.5\linewidth]{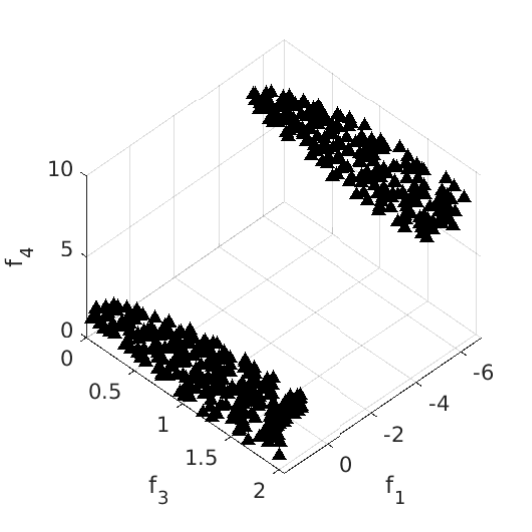}}
  \\
  \subfloat[$F(P_1)$]{%
       \includegraphics[width=0.5\linewidth]{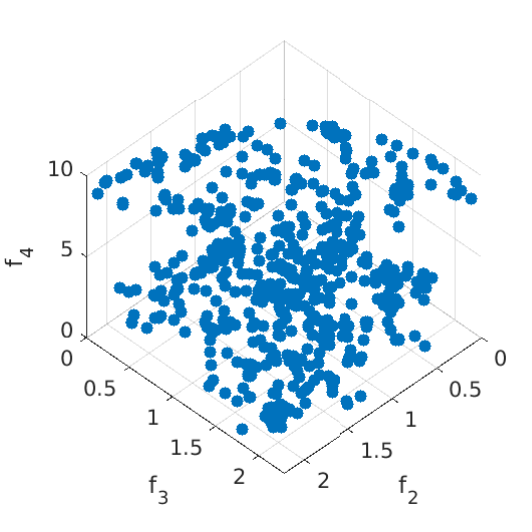}}
    \subfloat[Newton]{%
       \includegraphics[width=0.5\linewidth]{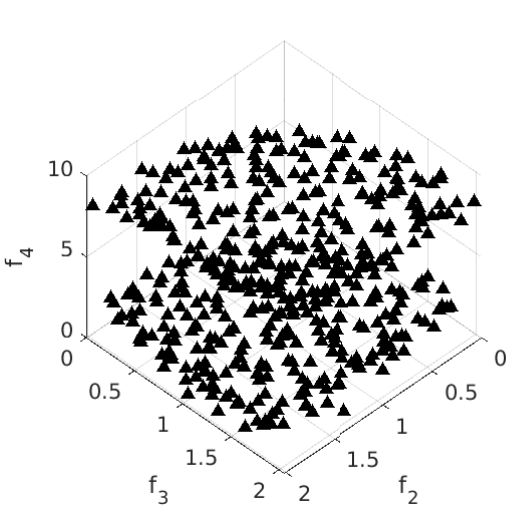}}
  \caption{Application of the $\Delta_p$-Newton method on CONV4-2F. The plots are the before-after results for all the projections not included in Figure \ref{fig:CONV4-2F_EA_123}.}
  \label{fig:CONV4-2F_EA_projections} 
\end{figure}

Figures \ref{fig:ZDT1_EA}-\ref{fig:CONV4-2F_EA_123} show the application of the Newton method with all the intermediate steps for one run of ZDT1, ZDT3, DTLZ1, DTLZ7, and one projection of CONV4-2F (defined below). Figure \ref{fig:CONV4-2F_EA_projections} shows the before-after results on the remaining projections of CONV4-2F. Figures \ref{fig:ZDT1_EA}-\ref{fig:CONV4-2F_EA_123} (a) and (b) show two of the populations obtained by NSGA-II for DTLZ1, ZDT1 and ZDT3 and NSGA-III for DTLZ7 and CONV4-2F. Figures \ref{fig:ZDT1_EA}-\ref{fig:CONV4-2F_EA_123} (c) show the filled set $Y$ and the selected $\mu$ elements used as reference $T$. Figures \ref{fig:ZDT1_EA}-\ref{fig:CONV4-2F_EA_123} (d) show the starting set for the Newton method $\X_0$ and the shifted reference $Z$. Finally, Figures \ref{fig:ZDT1_EA}-\ref{fig:CONV4-2F_EA_123} (e) and (f) show the Newton method's result and the performance, respectively. 

\begin{flalign*}
    \textbf{CON}&\textbf{V4-2F}\\
  F &:[-3,3]^4 \xrightarrow{} \mathbb{R}^4 \\
 F(x) &=  (f_1(x),f_2(x),f_3(x),f_4(x))^T \text{ where}:\\
 f_i(x) &=  \begin{cases}
    \|x+\mathds{1}-a^i\|^2 - 3.5 \sigma &\quad\text{if } x < (0,0,0,0)\\
    \|x-a^i\|^2 &\quad\text{otherwise}
  \end{cases}\\
  \phi_1 &= (0,\|a^1-a^2\|^2,\|a^1-a^3\|^2,\|a^1-a^4\|^2)\\
  \phi_4 &= (\|a^4-a^1\|^2,\|a^4-a^2\|^2,\|a^4-a^3\|^2,0)\\
  \sigma &= \phi_4-\phi_1\\
  a^1 &= (1,0,0,0), \quad a^2 = (0,1,0,0) \quad a^3 = (0,0,1,0),\\
  a^4 &= (0,0,0,1), \quad \mathds{1} = (1,1,1,1)\\
\end{flalign*}

\begin{figure} 
    \centering\captionsetup{width=.153\linewidth}
  \subfloat[CF2 Before]{%
       \includegraphics[width=0.20\linewidth]{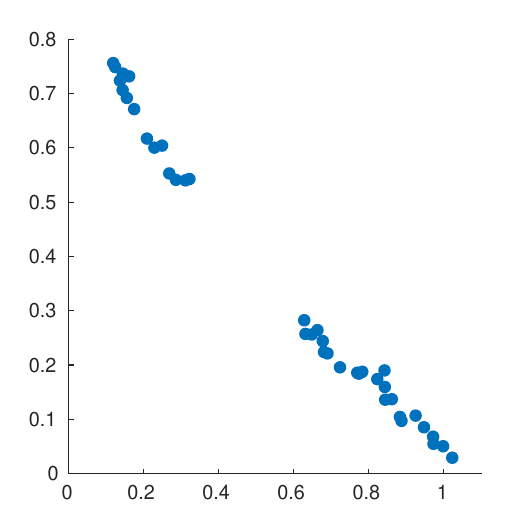}}
  \subfloat[ZDT3 Before]{%
        \includegraphics[width=0.20\linewidth]{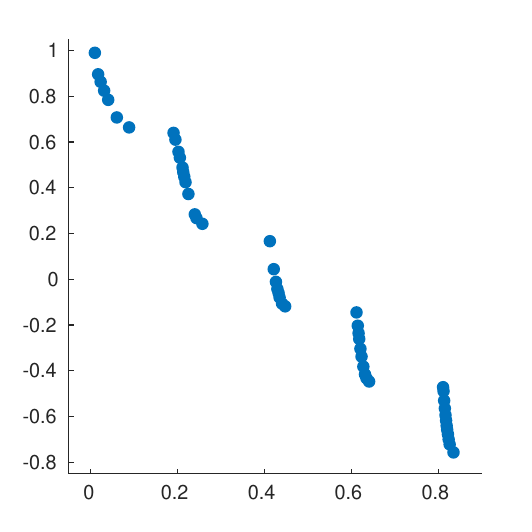}}
  \subfloat[DTLZ1 Before (NSGA-II)]{%
       \includegraphics[width=0.20\linewidth]{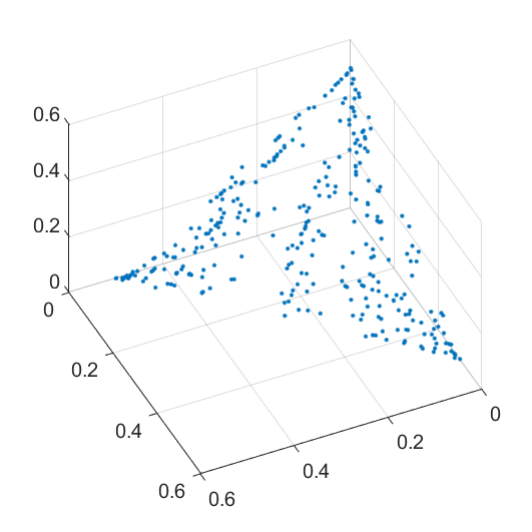}}
  \subfloat[DTLZ1 Before (NSGA-III)]{%
       \includegraphics[width=0.20\linewidth]{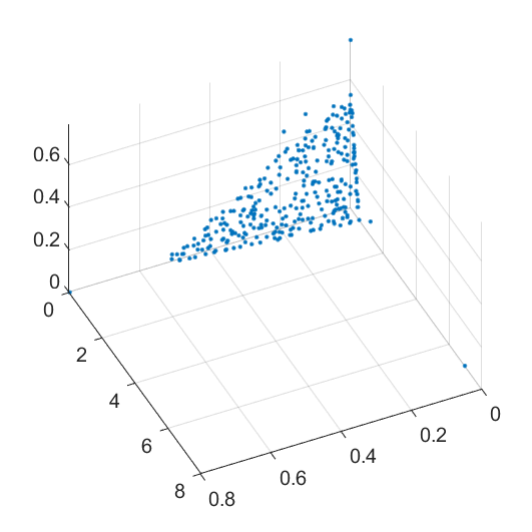}}
  \subfloat[DTLZ7 Before]{%
        \includegraphics[width=0.20\linewidth]{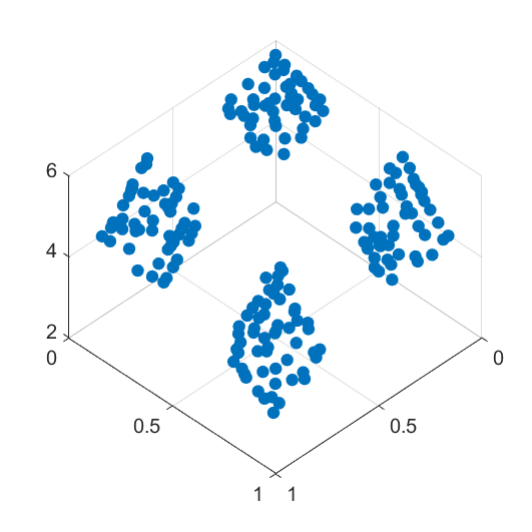}}
  \\
  \subfloat[CF2 After]{%
        \includegraphics[width=0.20\linewidth]{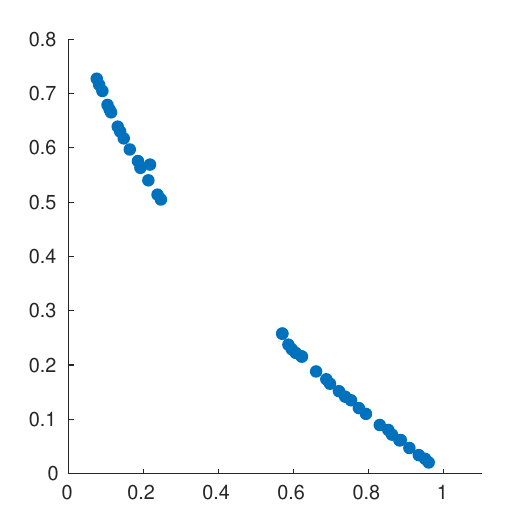}}
  \subfloat[ZDT3 After]{%
        \includegraphics[width=0.20\linewidth]{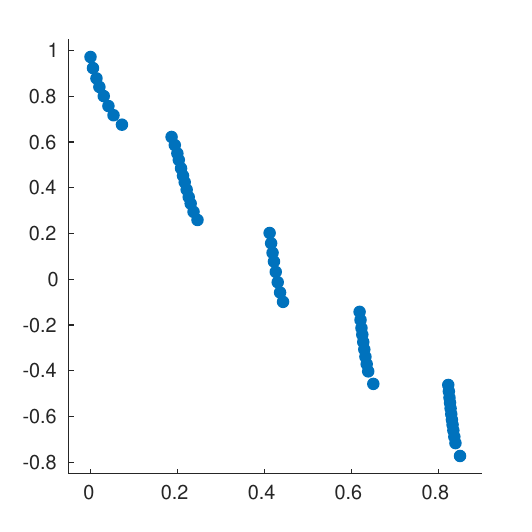}}
  \subfloat[DTLZ1 After (NSGA-II)]{%
       \includegraphics[width=0.20\linewidth]{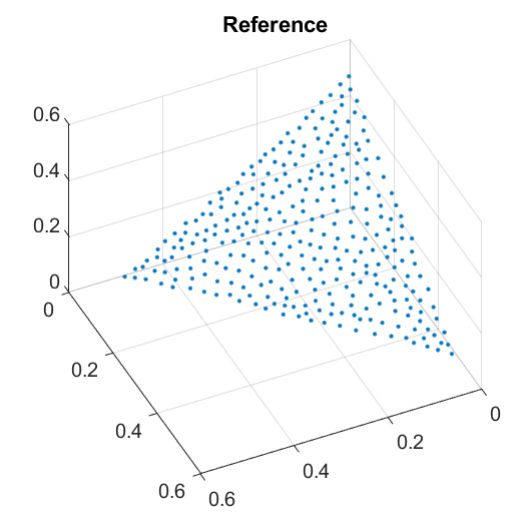}}
  \subfloat[DTLZ1 After (NSGA-III)]{%
       \includegraphics[width=0.20\linewidth]{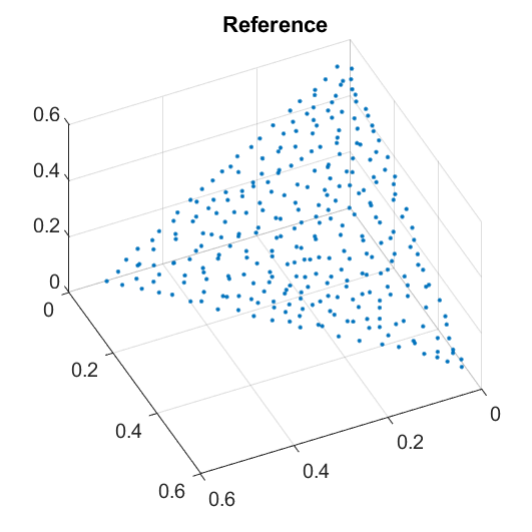}}
  \subfloat[DTLZ7 After]{%
        \includegraphics[width=0.20\linewidth]{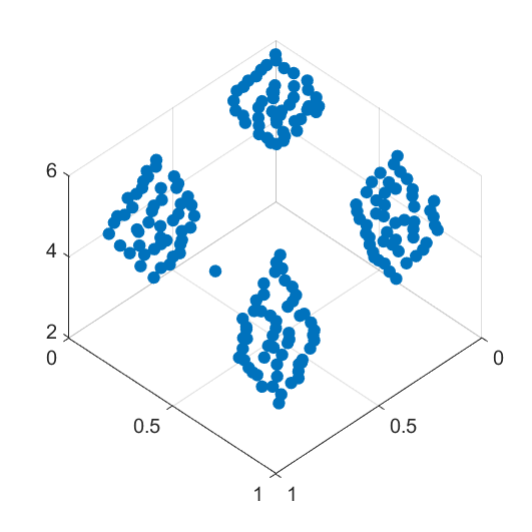}}
  \caption{Before and after Newton for CF2, ZDT3, CONV3 and DTLZ7.}
  \label{fig:before_after} 
\end{figure}

\subsection{Effect of Filling}

\begin{figure} 
    \centering\captionsetup{width=.45\linewidth}
  \subfloat[DTLZ1 without filling]{%
       \includegraphics[width=0.5\linewidth]{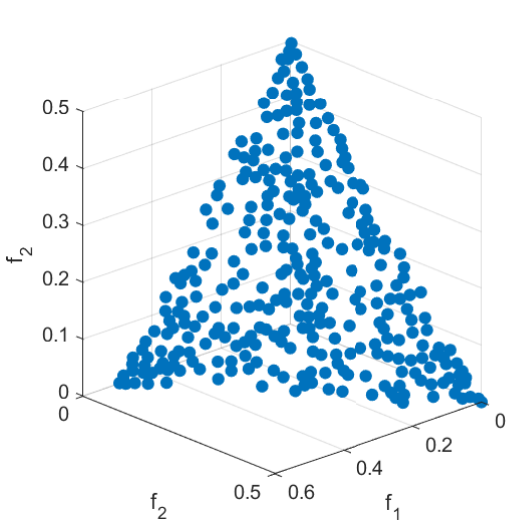}
       \label{subfig:DTLZ1nofill}}
    \subfloat[DTLZ1 with filling]{%
       \includegraphics[width=0.5\linewidth]{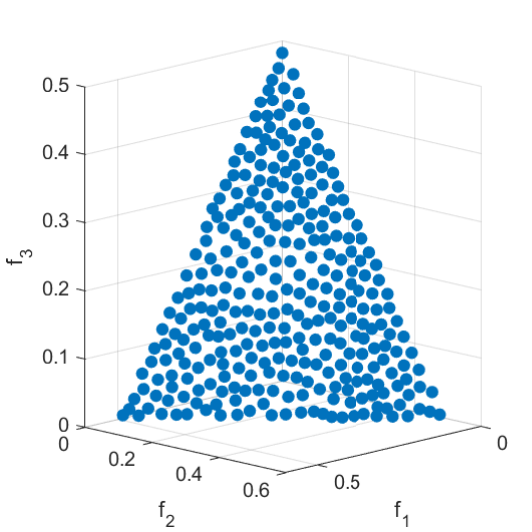}
       \label{subfig:DTLZ1fill}}
  \\
  \subfloat[DTLZ2 without filling]{%
       \includegraphics[width=0.5\linewidth]{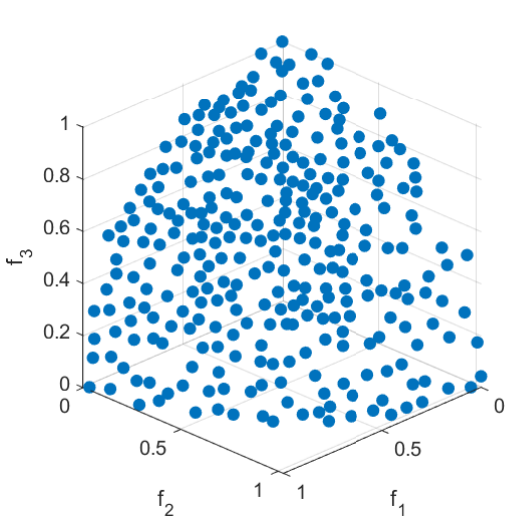}
       \label{subfig:DTLZ2nofill}}
    \subfloat[DTLZ2 with filling]{%
       \includegraphics[width=0.5\linewidth]{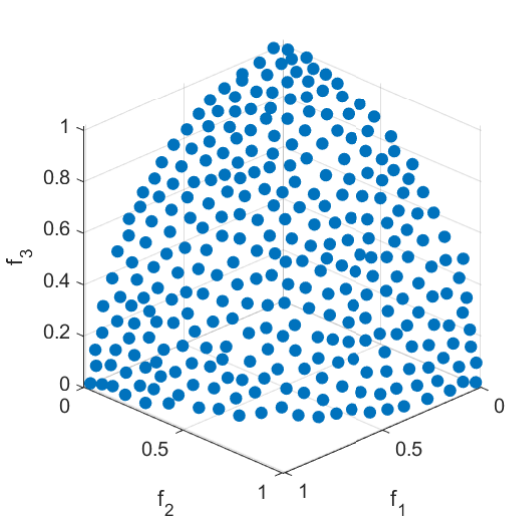}
       \label{subfig:DTLZ2fill}}
  \\
  \subfloat[ZDT4 without filling]{%
       \includegraphics[width=0.5\linewidth]{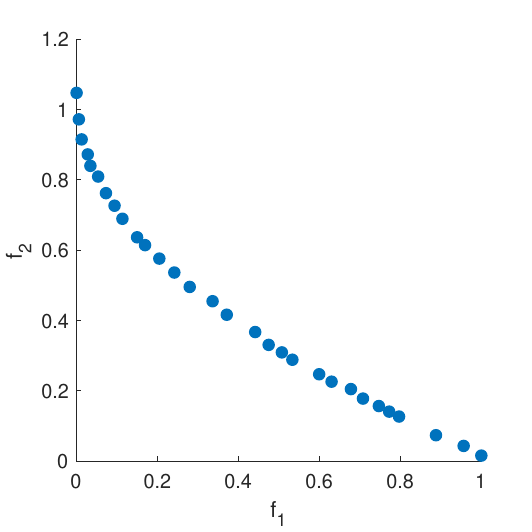}
       \label{subfig:ZDT4nofill}}
    \subfloat[ZDT4 with filling]{%
       \includegraphics[width=0.5\linewidth]{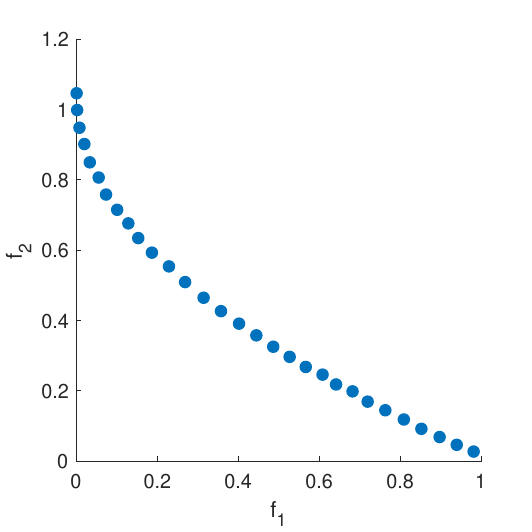}
       \label{subfig:ZDT4fill}}
  \\
  \caption{Effect of the filling on generating the reference set $Z$ on DTLZ1, DTLZ2 and ZDT4.}
  \label{fig:FillingEffect} 
\end{figure}

In this section, we motivate the use of the filling to generate the reference set $Z$ on the three examples shown in Figure \ref{fig:FillingEffect}. 
Figures \ref{subfig:DTLZ1nofill} and \ref{subfig:DTLZ1fill} show the generated 
reference set $Z$ with and without filling, respectively, for DTLZ1. Both sets
have been generated using the same data from an NSGA-II run. 
Figures \ref{subfig:DTLZ2nofill} and \ref{subfig:DTLZ2fill} show analog results
for DTLZ2 using SMS-EMOA populations. Finally,  Figures \ref{subfig:DTLZ2nofill} and \ref{subfig:DTLZ2fill} are results for ZDT4 and NSGA-II. In all cases, the 
reference set is more evenly spread around the Pareto front when using the filling step. 

\end{document}

%% file: results-gen=300.tex
\begin{table*}[t]
\label{tab:numerical-results}
\caption{Warmstarting the \algo method after 300 iterations of the MOEA, we show the $\Delta_2$ values 
(median and 10\% - 90\% quantile range) of the final Pareto fronts, averaged over
30 independent runs. The Newton method is executed for six iterations, and the MOEA terminates at comparable function evaluations.
Mann–Whitney U test (with 5\% significance level) is employed to compare the performance of the Newton method and the MOEA: $\uparrow$ (\algo surpasses MOEA), $\leftrightarrow$ (no significant difference), and $\downarrow$ (\algo loses) where the Holm-Sidak method is used to adjust the $p$-value for multiple testing. 
}

\begin{minipage}{0.5\textwidth}
\footnotesize
\begin{tabular}{llll}
\toprule
Method & Problem & MOEA + \algo  & MOEA \\
\midrule
NSGA-II & CF1 & 0.0300(4.463e-02)$\uparrow$ & 0.0833(1.110e-01) \\
NSGA-II & CF2 & 0.0886(9.911e-02)$\leftrightarrow$ & 0.0955(2.213e-01) \\
NSGA-II & CF3 & 0.3382(2.004e-01)$\uparrow$ & 0.4234(4.581e-01) \\
NSGA-II & CF4 & 0.1457(1.513e-01)$\leftrightarrow$ & 0.1281(2.713e-01) \\
NSGA-II & CF5 & 0.2882(4.215e-01)$\leftrightarrow$ & 0.3773(3.223e-01) \\
NSGA-II & CF6 & 0.2598(1.903e-01)$\leftrightarrow$ & 0.1993(1.654e-01) \\
NSGA-II & CF7 & 0.3080(3.629e-01)$\leftrightarrow$ & 0.3758(4.328e-01) \\
NSGA-II & CF8 & 0.7557(2.343e-01)$\uparrow$ & 1.1986(1.185e+00) \\
NSGA-II & CF9 & 0.1568(5.746e-02)$\uparrow$ & 0.8577(6.374e-01) \\
NSGA-II & CF10 & 0.7984(1.400e+00)$\uparrow$ & 10.4022(1.482e+01) \\
NSGA-II & ZDT1 & 0.0048(2.183e-04)$\uparrow$ & 0.0057(7.470e-04) \\
NSGA-II & ZDT2 & 0.0046(1.077e-04)$\uparrow$ & 0.0059(6.459e-04) \\
NSGA-II & ZDT3 & 0.0071(1.398e-03)$\leftrightarrow$ & 0.0068(8.095e-04) \\
NSGA-II & ZDT4 & 0.0048(2.578e-04)$\uparrow$ & 0.0054(5.038e-04) \\
NSGA-II & ZDT6 & 0.0047(5.787e-04)$\uparrow$ & 0.0049(6.355e-04) \\
NSGA-II & DTLZ1 & 0.0124(6.605e-04)$\uparrow$ & 0.0179(1.582e-03) \\
NSGA-II & DTLZ2 & 0.0429(1.694e-03)$\uparrow$ & 0.0460(3.702e-03) \\
NSGA-II & DTLZ3 & 0.0435(2.139e-03)$\uparrow$ & 0.0473(3.356e-02) \\
NSGA-II & DTLZ4 & 0.0438(1.562e-03)$\uparrow$ & 0.0448(3.131e-03) \\
NSGA-II & DTLZ5 & 0.0045(2.844e-04)$\uparrow$ & 0.0046(1.304e-04) \\
NSGA-II & DTLZ6 & 0.0431(4.569e-02)$\uparrow$ & 0.0573(4.226e-02) \\
NSGA-II & DTLZ7 & 0.0399(5.767e-03)$\uparrow$ & 0.0484(7.544e-03) \\
NSGA-II & IDTLZ1 & 0.0747(3.750e-03)$\uparrow$ & 0.0953(3.511e-03) \\
NSGA-II & IDTLZ2 & 1.1248(2.553e-02)$\uparrow$ & 1.1754(3.821e-02) \\
NSGA-II & IDTLZ3 & 950.0772(1.188e+01)$\uparrow$ & 983.7051(3.063e+01) \\
NSGA-II & IDTLZ4 & 1.7877(4.016e-02)$\uparrow$ & 1.8510(5.479e-02) \\
NSGA-II & CONV4-2F & 0.4360(2.821e-02)$\uparrow$ & 0.4790(8.461e-02) \\
NSGA-III & CF1 & 0.0226(5.467e-03)$\uparrow$ & 0.1866(2.179e-01) \\
NSGA-III & CF2 & 0.0792(4.738e-02)$\uparrow$ & 0.3172(4.645e-01) \\
NSGA-III & CF3 & 0.2862(1.860e-01)$\uparrow$ & 0.6304(6.208e-01) \\
NSGA-III & CF4 & 0.1200(4.080e-02)$\uparrow$ & 0.2463(9.613e-01) \\
NSGA-III & CF5 & 0.4874(4.474e-01)$\uparrow$ & 0.5778(7.658e-01) \\
NSGA-III & CF6 & 0.2409(1.911e-01)$\leftrightarrow$ & 0.1777(1.621e-01) \\
NSGA-III & CF7 & 0.3748(3.103e-01)$\uparrow$ & 0.5500(8.484e-01) \\
NSGA-III & CF8 & 0.7942(1.870e-01)$\uparrow$ & 1.1450(1.050e+00) \\
NSGA-III & CF9 & 0.1870(9.110e-02)$\uparrow$ & 0.4705(1.778e+00) \\
NSGA-III & CF10 & 0.7744(8.865e-01)$\leftrightarrow$ & 1.4668(9.989e+00) \\
NSGA-III & ZDT1 & 0.0057(2.432e-03)$\leftrightarrow$ & 0.0064(2.935e-05) \\
NSGA-III & ZDT2 & 0.0047(9.501e-05)$\uparrow$ & 0.0055(3.772e-05) \\
NSGA-III & ZDT3 & 0.0064(2.571e-03)$\uparrow$ & 0.0102(1.880e-04) \\
NSGA-III & ZDT4 & 0.0066(1.070e-02)$\uparrow$ & 0.0094(7.426e-02) \\
NSGA-III & ZDT6 & 0.0037(7.453e-05)$\uparrow$ & 0.0058(9.301e-04) \\
NSGA-III & DTLZ1 & 0.0127(3.195e-04)$\downarrow$ & 0.0112(5.865e-05) \\
NSGA-III & DTLZ2 & 0.0337(7.155e-04)$\downarrow$ & 0.0302(9.015e-05) \\
NSGA-III & DTLZ3 & 0.0338(6.955e-04)$\downarrow$ & 0.0304(1.033e-03) \\
NSGA-III & DTLZ4 & 0.0339(8.853e-04)$\downarrow$ & 0.0302(4.015e-05) \\
NSGA-III & DTLZ5 & 0.0046(3.616e-04)$\uparrow$ & 0.0250(6.557e-04) \\
NSGA-III & DTLZ6 & 0.0368(2.585e-02)$\uparrow$ & 0.0869(1.319e-01) \\
NSGA-III & DTLZ7 & 0.0407(5.684e-03)$\uparrow$ & 0.0588(1.104e-03) \\
\end{tabular}
\end{minipage}
\hfill
\begin{minipage}{0.5\textwidth}
\footnotesize
\begin{tabular}{llll}
NSGA-III & IDTLZ1 & 0.0758(3.981e-03)$\uparrow$ & 0.0890(1.333e-03) \\
NSGA-III & IDTLZ2 & 1.0408(2.159e-02)$\uparrow$ & 1.0769(6.307e-03) \\
NSGA-III & IDTLZ3 & 930.5973(3.303e+01)$\downarrow$ & 922.9865(1.259e+01) \\
NSGA-III & IDTLZ4 & 1.6451(6.316e-02)$\uparrow$ & 1.7512(5.912e-03) \\
NSGA-III & CONV4-2F & 0.4414(3.165e-02)$\uparrow$ & 0.5209(1.176e-01) \\
SMS-EMOA & CF1 & 0.0244(1.041e-02)$\uparrow$ & 0.0799(2.054e-01) \\
SMS-EMOA & CF2 & 0.1571(1.121e-01)$\leftrightarrow$ & 0.1285(2.332e-01) \\
SMS-EMOA & CF3 & 0.3240(1.063e-01)$\uparrow$ & 0.4563(3.103e-01) \\
SMS-EMOA & CF4 & 0.1262(1.531e-01)$\leftrightarrow$ & 0.1318(1.632e-01) \\
SMS-EMOA & CF5 & 0.5008(4.483e-01)$\leftrightarrow$ & 0.4526(4.302e-01) \\
SMS-EMOA & CF6 & 0.2556(1.042e-01)$\downarrow$ & 0.2081(1.691e-01) \\
SMS-EMOA & CF7 & 0.4123(4.104e-01)$\leftrightarrow$ & 0.4588(4.637e-01) \\
SMS-EMOA & CF8 & 0.7417(1.973e-01)$\uparrow$ & 3.3526(2.388e+00) \\
SMS-EMOA & CF9 & 0.5010(1.408e-01)$\downarrow$ & 0.3871(2.644e-01) \\
SMS-EMOA & CF10 & 0.5504(6.736e-01)$\uparrow$ & 1.1320(3.102e+00) \\
SMS-EMOA & ZDT1 & 0.0045(1.414e-04)$\downarrow$ & 0.0043(6.758e-05) \\
SMS-EMOA & ZDT2 & 0.0048(1.206e-04)$\uparrow$ & 0.0059(8.305e-04) \\
SMS-EMOA & ZDT3 & 0.0062(4.074e-04)$\leftrightarrow$ & 0.0062(5.672e-04) \\
SMS-EMOA & ZDT4 & 0.0046(4.369e-04)$\leftrightarrow$ & 0.0045(6.858e-03) \\
SMS-EMOA & ZDT6 & 0.0037(9.603e-05)$\uparrow$ & 0.0041(3.348e-04) \\
SMS-EMOA & DTLZ1 & 0.0128(1.279e-03)$\uparrow$ & 0.0295(1.020e-01) \\
SMS-EMOA & DTLZ2 & 0.0336(1.008e-03)$\uparrow$ & 0.0627(3.314e-02) \\
SMS-EMOA & DTLZ3 & 0.0342(3.620e-03)$\uparrow$ & 0.0720(4.076e-01) \\
SMS-EMOA & DTLZ4 & 0.0341(8.198e-04)$\uparrow$ & 0.0504(7.836e-03) \\
SMS-EMOA & DTLZ5 & 0.0055(3.670e-03)$\uparrow$ & 0.0071(6.424e-03) \\
SMS-EMOA & DTLZ6 & 0.0481(5.194e-02)$\leftrightarrow$ & 0.0470(6.263e-02) \\
SMS-EMOA & DTLZ7 & 0.0523(4.408e-02)$\leftrightarrow$ & 0.0509(2.074e-02) \\
SMS-EMOA & IDTLZ1 & 0.0744(2.376e-03)$\uparrow$ & 0.0882(2.154e-02) \\
SMS-EMOA & IDTLZ2 & 0.7901(5.179e-02)$\leftrightarrow$ & 0.7969(9.259e-02) \\
SMS-EMOA & IDTLZ3 & 757.5839(7.299e+01)$\downarrow$ & 659.7112(5.920e+01) \\
SMS-EMOA & IDTLZ4 & 1.7023(5.528e-02)$\leftrightarrow$ & 1.7127(1.212e-01) \\
SMS-EMOA & CONV4-2F & 0.5275(2.942e-01)$\uparrow$ & 0.7978(4.228e-01) \\
MOEAD & ZDT1 & 0.0073(3.062e-03)$\uparrow$ & 0.0158(7.974e-02) \\
MOEAD & ZDT2 & 0.0056(5.081e-04)$\uparrow$ & 0.0589(2.719e-01) \\
MOEAD & ZDT3 & 0.0497(1.163e-01)$\leftrightarrow$ & 0.0432(1.417e-01) \\
MOEAD & ZDT4 & 0.0206(2.553e-02)$\uparrow$ & 2.0001(4.178e+00) \\
MOEAD & ZDT6 & 0.0062(1.078e-03)$\uparrow$ & 0.2462(3.517e-01) \\
MOEAD & DTLZ1 & 0.0123(2.846e-04)$\uparrow$ & 0.0132(3.123e-04) \\
MOEAD & DTLZ2 & 0.0339(7.752e-04)$\uparrow$ & 0.0346(2.125e-05) \\
MOEAD & DTLZ3 & 0.0338(8.183e-04)$\uparrow$ & 0.0406(1.650e-02) \\
MOEAD & DTLZ4 & 0.0339(9.495e-04)$\uparrow$ & 0.0346(6.397e-01) \\
MOEAD & DTLZ5 & 0.1649(1.460e-02)$\downarrow$ & 0.0249(1.428e-03) \\
MOEAD & DTLZ6 & 0.0571(7.161e-02)$\uparrow$ & 0.0883(1.621e-01) \\
MOEAD & DTLZ7 & 0.0912(3.172e-02)$\leftrightarrow$ & 0.0939(7.184e-03) \\
MOEAD & IDTLZ1 & 0.0131(1.155e-03)$\uparrow$ & 0.0206(7.493e-04) \\
MOEAD & IDTLZ2 & 1.0747(3.208e-02)$\uparrow$ & 1.1985(7.294e-03) \\
MOEAD & IDTLZ3 & 930.6047(1.901e+01)$\uparrow$ & 999.5192(1.280e+01) \\
MOEAD & IDTLZ4 & 1.7920(3.396e-02)$\uparrow$ & 1.8660(5.549e-03) \\
MOEAD & CONV4-2F & 1.1570(4.489e-02)$\downarrow$ & 0.4524(1.109e-01) \\
\hline
+/$\approx$/- &  & 66/21/11 &  \\
\bottomrule
\end{tabular}
\end{minipage}
\end{table*}